\documentclass[letterpaper]{article} 
\usepackage[preprint]{aaai2027}  
\usepackage[hyphens]{url}  
\usepackage{graphicx} 
\usepackage{natbib}  
\usepackage{caption} 
\usepackage{algorithm}
\usepackage{algorithmic}

\usepackage{newfloat}
\usepackage{listings}
\DeclareCaptionStyle{ruled}{labelfont=normalfont,labelsep=colon,strut=off} 
\floatstyle{ruled}
\newfloat{listing}{tb}{lst}{}
\floatname{listing}{Listing}
\usepackage{multirow}
\usepackage{amsmath, amssymb} 
\usepackage{ulem}     
\usepackage{graphicx}    
\usepackage{natbib}          
\usepackage{booktabs}

\title{Efficient RLVR Scheduling via Graph-Structured Online Difficulty Estimation}
\author{
    Zhizhao Liu,
    Zhiliang Tian,
    Xi Wang,
    Zhihua Wen,
    Yihang Xiong,
    Zhiquan Lai,
    Dongsheng Li,
}
\affiliations{
    \textsuperscript{\rm 1}PDL Lab, College of Computer Science and Technology, National University of Defense Technology\\

    No.109 Deya Road, Kaifu District, Changsha\\
    Hunan 410073, P.R. China\\
    tianzhiliang@nudt.edu.cn\\
    dsli@nudt.edu.cn
}

\begin{document}

\maketitle

\begin{abstract}
Reinforcement learning with verifiable rewards (RLVR) improves the reasoning capabilities of large language models but relies on costly rollout exploration. Assigning the same exploration budget to samples with different difficulty levels is inefficient: easy samples may receive redundant rollouts, whereas difficult but learnable samples may receive too little exploration. Existing adaptive schedulers address this mismatch through curriculum-based sample selection or non-uniform rollout allocation based on estimated sample difficulty. However, obtaining reliable online difficulty estimates remains challenging: dedicated probing adds substantial generation overhead, whereas history-based estimators face a cold start with no initial observations and stale feedback, and typically ignore relations among samples. To address these limitations, we propose a plug-and-play graph-based online difficulty estimator that shares rollout feedback across related samples and continuously updates their difficulty estimates, mitigating cold start and staleness without dedicated probing. Specifically, we first construct a difficulty-aware sample graph based on semantic and reasoning similarities. Based on this graph, we introduce latent difficulty states and use a Potts prior to encourage neighboring samples to share the same state. We then employ a state-level Beta-Binomial model to aggregate the rollout outcomes associated with each state. Finally, we use an online mean-field variational algorithm to continuously update the latent-state assignments and state-level difficulty as new feedback arrives. Our framework can be integrated into sample-selection and rollout-allocation schedulers, enabling difficulty-adaptive exploration without dedicated probing. Experiments across multiple base models, RL schedulers, and benchmarks demonstrate that our framework achieves better performance under matched rollout budgets.

\end{abstract}

\section{Introduction}

Reinforcement learning with verifiable rewards (RLVR) improves the reasoning capabilities of large language models through automatically verifiable feedback~\cite{shao2024deepseekmath,yu2026dapo,zheng2025groupgspo}. Group-based algorithms, such as Group Relative Policy Optimization (GRPO), generate multiple responses for each training sample and compare their rewards to update the policy~\cite{shao2024deepseekmath}. These rollouts explore alternative solutions but also incur substantial generation cost~\cite{guo2025deepseekr1,yao2026optimizinggvm}. 

However, GRPO commonly assigns the same number of rollouts to every selected sample~\cite{shao2024deepseekmath,yao2026optimizinggvm}. This uniform allocation treats all samples as if they require the same amount of exploration, although the current policy may easily solve some samples and struggle with others. Additional rollouts on easy samples often repeat predictable outcomes, whereas difficult but learnable samples may receive too few attempts to discover useful solution trajectories~\cite{nguyen2026adaptivevip,yao2026optimizinggvm,zeng2026cures}. This budget mismatch motivates difficulty-aware scheduling, which uses sample difficulty to decide how to distribute the available exploration budget.

Existing studies implement difficulty-aware scheduling in two ways. First, sample-selection methods adjust the training distribution and decide which samples receive exploration~\cite{pcl,qu2026can,cdas}. Second, rollout-allocation methods assign different numbers of rollouts to selected samples and determine how much exploration each sample receives~\cite{nguyen2026adaptivevip,yao2026optimizinggvm,zeng2026cures}. Although these methods make different scheduling decisions, both need up-to-date difficulty estimates because the policy continually changes which samples it can solve. This requirement makes low-cost online difficulty estimation essential for adaptive RLVR scheduling.

However, obtaining such up-to-date and reliable estimates remains challenging. 
Some researchers explicitly estimate success probabilities for train examples under the current policy by using additional rollout probing~\cite{zeng2026cures,yao2026optimizinggvm,team2025kimi}. Although this approach provides estimation, it incurs substantial computational overhead and introduces statistical uncertainty due to the limited number of stochastic rollouts\footnote{An intuitive analogy: tossing a fair coin 4 times still has a 12.5\% chance of coming up all heads or all tails.}. Further, some researchers estimate these success probabilities from outcomes collected during previous training steps~\cite{nguyen2026adaptivevip,nguyen2026space,cdas,qu2026can}. Although this approach reduces the additional rollout overhead, it struggles with cold start and accuracy because historical results may introduce noise from the variance of stochastic outcomes, and the historical results may become outdated as the policy model evolves during training.
Thus, these limitations motivate the following question:
\textit{How can we continuously estimate the evolving difficulty of all training examples at low cost as the policy evolves during RLVR training?}

We argue that semantic or reasoning structure related samples can provide mutually informative evidence for difficulty estimation. Samples with similar semantic often rely on similar model capabilities and may exhibit exploitable local statistical correlations. Thus, in this paper, we propose a graph-based online difficulty estimation framework for RLVR-based LLM training, which can be integrated into different RL training schedulers to improve their performance. Specifically, we first construct a difficulty-aware sample graph that connects samples with similar semantic and reasoning characteristics. Then, we model sample difficulty as graph-structured latent states and develop an online variational inference algorithm to update these states using rollout feedback collected during training. In this way, related samples can share historical evidence, enabling us to continuously estimate the success probability of all training examples without additional rollout probing. Finally, we integrate our estimator into sample selection or rollout allocation methods to improve RLVR training efficiency. Experiments across different models and datasets demonstrate the accuracy, efficiency, and generality of our method.

Our contributions are as follows: (1) We are the first to formulate dynamic difficulty estimation in RLVR as a graph-structured latent-variable inference problem. (2) We propose a low-cost online estimator that propagates rollout feedback among related samples and continuously estimates their success probabilities as the policy evolves. (3) We integrate our estimator into both sample selection and rollout allocation methods. Extensive experiments show that our method is cold start friendly and substantially reduces difficulty estimation overhead while consistently improving the downstream performance of RLVR schedulers.
\section{Related Work}
\begin{figure*}[h]
    \centering
    \includegraphics[width=1\linewidth]{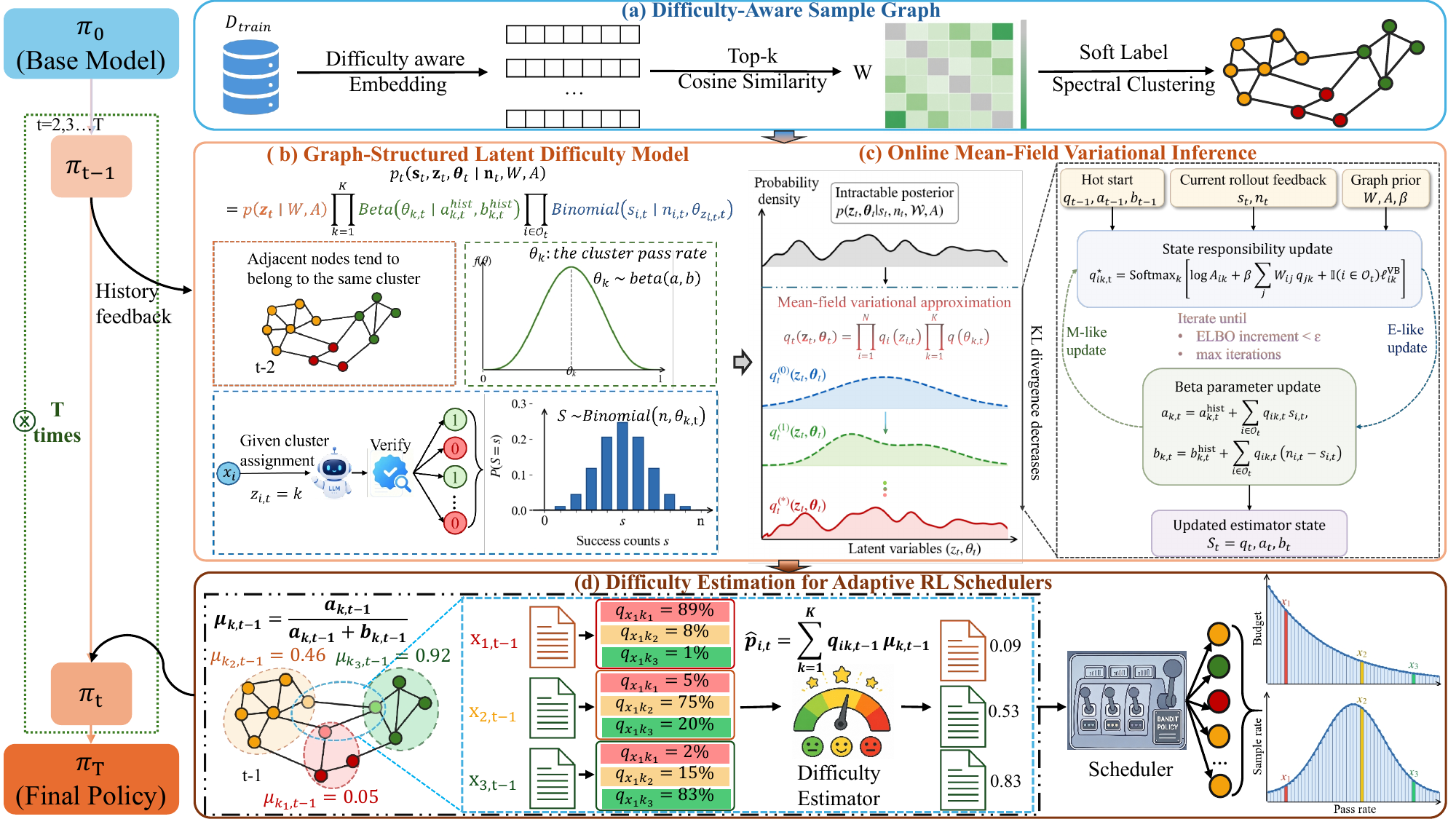}
    \caption{The overview of our framework. We first construct a difficulty-aware sample graph (a). During RL training, our graph-structured latent model aggregates rollout feedback across related samples (b), while online mean-field variational inference updates their evolving difficulty (c). The resulting estimates guide adaptive sample selection and rollout allocation (d).}
    \label{fig:main}
\end{figure*}

\subsection{Adaptive Sampling and Rollout Allocation in RLVR}

Curriculum learning order or reweight examples according to difficulty, competence, or learnability~\cite{bengio2009icml-curriculum,pmlr-v162-mindermann22a}. This principle is especially relevant to group-based RLVR, where prompts with nearly deterministic group rewards provide little group-relative learning signal~\cite{shao2024deepseekmath,yu2026dapo,zheng2025groupgspo}. Prompt-selection methods therefore use estimated difficulty, competence alignment, uncertainty, or reward dynamics to prioritize informative examples~\cite{yu2026dapo,pcl,qu2026can,cdas,zheng2026actgreso,bae2026online}. A complementary line allocates non-uniform rollout counts. GVM, VIP, and CurES use gradient-variance or stability objectives~\cite{yao2026optimizinggvm,nguyen2026adaptivevip,zeng2026cures}, whereas recent methods use sequential uncertainty reduction, Bernoulli-variance proxies, or posterior hit utility~\cite{xiong2025reinforce,fang2026allocate}. Although their scheduling objectives differ, these methods require estimates of current prompt behavior. Our work is complementary: it provides a reusable estimator rather than another scheduling rule.

\subsection{Dynamic Difficulty Estimation for Reasoning LLMs}
Difficulty depends not only on an example but also on the policy model. Item response theory explicitly models the interaction between latent ability and item difficulty and has been adapted to NLP evaluation~\cite{lalor-etal-2016-building,rodriguez-etal-2021-evaluation}. Training-dynamics methods instead characterize examples through quantities such as confidence and variability across epochs, or prioritize examples that remain learnable and useful~\cite{swayamdipta-etal-2020-dataset,pmlr-v162-mindermann22a}. RLVR further requires these model-dependent estimates to track an evolving policy.
Existing RLVR estimators follow three broad strategies. Model-based methods learn a value or scoring model to predict difficulty without rolling out every candidate~\cite{pcl}. Probing-based methods estimate current success probabilities from additional on-policy generations, obtaining timely but costly and noisy signals~\cite{yao2026optimizinggvm,zeng2026cures,team2025kimi}. History-based methods reuse training feedback: MoPPS performs streaming Bayesian inference~\cite{qu2026can}; CDAS aggregates historical performance discrepancies~\cite{cdas}; GRESO exploits temporal regularity in rewards~\cite{zheng2026actgreso}; and VIP fits a lightweight predictor to recent rollouts~\cite{nguyen2026adaptivevip}. Complementary scheduling methods operate at different granularities: Re-Schedule derives a static structural-learnability score from offline reasoning trees for easy-to-hard curriculum scheduling~\cite{wang2026scheduling}, whereas ARRoL predicts final correctness from partial rollouts and prunes trajectories online to balance binary rewards~\cite{xu2026prunegenerateonlinerollout}. Learned predictors must remain calibrated under policy drift, probing and tree construction incur generation overhead, and historical observations may be sparse or stale. 
Our approach also draws on graph-based semi-supervised learning and probabilistic graphical models, using spectral structure, Potts priors, and mean-field inference to propagate sparse feedback across related samples; see Appendix H for further discussion.
In contrast, our estimator couples related samples through an explicit graph, pools sparse evidence through shared latent difficulty states.

Our setting combines these ideas with policy-induced nonstationarity: rollout observations are sparse, while their generating policy changes throughout training. We construct a difficulty-aware graph, associate prompts with time-indexed latent states, aggregate binary outcomes through state-level Beta--Binomial factors, and carry the variational posterior across training steps. This yields sequential probabilistic reasoning over relational and evolving latent states, together with the success-probability estimates required by adaptive RLVR schedulers.
\section{Method}

\subsection{3.1 Overview}

We propose a graph-based online framework for estimating the evolving difficulty of training examples in RLVR. By leveraging historical rollout outcomes from related examples, our framework predicts the success probability and difficulty of each example under the current policy before each training step.
As illustrated in Fig.~\ref{fig:main}, the framework consists of four sequential modules. First, the \textbf{Difficulty-aware Sample Graph} jointly encodes the semantic and difficulty features of training examples and constructs a sparse train set graph (Sec.~3.3). Second, the \textbf{Graph-Structured Latent Difficulty Model} maps the graph into latent difficulty states that can be dynamically updated by rollouts feedback, with examples in the same state sharing a common success probability (Sec.~3.4). Third, the \textbf{Online Mean-Field Variational Inference} updates the latent-state distribution of each example and the posterior success probability of each latent state (Sec.~3.5). Finally, we use the updated posterior to estimate the success probability and difficulty of candidate samples for the next training step (Sec.~3.6).

\subsection{3.2 Problem Formulation}
Let the training dataset for reinforcement learning be $\mathcal{D}=\{x_i\}_{i=1}^{N}$, where $x_i$ denotes the $i$-th sample. Following the actual training order, we let $\mathcal{O}_t$ represent the set of samples for which we generate rollouts at training step $t$. Under the current policy model $\pi_t$, each sample $i\in\mathcal{O}_t$ generates $n_{i,t}$ rollouts, and we denote the binary correctness outcome of the $g$-th rollout by $y_{i,t,g}\in\{0,1\}$. We aggregate the rollout outcomes of the same sample into the number of successful rollouts, $s_{i,t}=\sum_{g=1}^{n_{i,t}}y_{i,t,g}$, and define its empirical success rate as $r_{i,t}=s_{i,t}/n_{i,t}$.

Our goal is to accurately estimate the probability that the current policy correctly solves each sample before each RL training step: $\hat p_{i,t} = \Pr_{Y\sim\pi_t}(Y\text{ is correct} \mid x_i, \mathbf{s}_{<t}, \mathbf{n}_{<t})$. We then define the difficulty of sample $i$ at step $t$ as $d_{i,t} = 1 - p_{i,t}$. Finally, we feed these probability or difficulty estimates into existing RL scheduling frameworks to guide decisions such as sample selection and rollout allocation.

\subsection{3.3 Difficulty-Aware Sample Graph}
In this section, we construct a sparse similarity graph over the training dataset based on semantic and difficulty. This allows samples with few or no prior rollouts to estimate their current expected difficulty by leveraging historical feedback from semantically and difficulty-wise similar neighbors.

To encourage the encoder to capture semantic content, reasoning structure, and difficulty-related characteristics of each sample, we introduce a difficulty-aware instruction $I_{\pi}$:
\textit{Instruction: Given a prompt, encode it into a dense vector that captures both its semantic content and difficulty level, for the purpose of similarity comparison. \textbackslash n Prompt: [SAMPLE]}
Let $f_{\mathrm{emb}}$ denote a pretrained embedding model. The representation of sample $x_i$ is computed as $\widetilde{\boldsymbol{\phi}}_i = f_{\mathrm{emb}}(I_{\pi}, x_i)$ and normalized to $\boldsymbol{\phi}_i = \widetilde{\boldsymbol{\phi}}_i / \|\widetilde{\boldsymbol{\phi}}_i\|_2$. For each pair of samples, we measure their similarity with cosine similarity: $c_{ij} = \boldsymbol{\phi}_i^\top \boldsymbol{\phi}_j$.

To capture the underlying manifold structure of high-dimensional sample embeddings while reducing computational cost, we construct a $k$-nearest-neighbor graph by retaining the $k_{\mathrm{nn}}$ nearest neighbors of each node. We then retain only edges where the two nodes are mutual $k$-nearest neighbors and remove those with negative similarities, yielding a sparse undirected sample graph $W \in \mathbb{R}^{N \times N}$. Specifically, $W_{ij} = c_{ij}$ if $x_i$ and $x_j$ are mutual $k$-nearest neighbors and $c_{ij} > 0$, and $0$ otherwise. A larger $W_{ij}$ reflects greater similarity in semantic content, reasoning structure, and difficulty-related features between $x_i$ and $x_j$.

\subsection{3.4 Graph-Structured Latent Difficulty Model}
In this section, we aim to perform statistical inference of sample difficulty by leveraging the graph structure and the historical rollout outcomes collected during previous RL training. Specifically, we first construct a graph-structured prior over sample latent states, then model rollout feedback through shared state-level success probabilities, and finally derive the posterior distribution for subsequent difficulty estimation.

\subsubsection{Graph Prior.} Although $W$ captures local similarity relationships among samples, it falls short in describing the overall difficulty structure of the training dataset. To further uncover similarities in both semantic features and difficulty levels across samples, we apply spectral clustering to the sparse undirected sample graph $W \in \mathbb{R}^{N \times N}$, and it yields an initial set of static cluster labels $c_i \in \{1,\ldots,K\}$.
However, spectral clustering relies solely on static sample representations, and its assignments can be affected by representation and clustering errors. More importantly, as we continuously update the policy during RL training, the difficulty of the same sample under the current policy may also change. 
Therefore, we convert each static cluster label into a smoothed prior distribution over latent states, we define $\displaystyle A_{ik}=\begin{cases}1-\epsilon, & k=c_i,\\\frac{\epsilon}{K-1}, & k\neq c_i,\end{cases}$. This conversion retains the structure from spectral clustering, and it also allows subsequent rollout outcomes to refine each sample's latent assignment.
Then, inspired by~\citet{besag1974spatialpotts}, we introduce a Potts graph prior~\cite{potts1952somepotts} over the latent assignments to retain the initial clustering structure while encouraging neighboring samples to share the same latent state. Specifically, we define
$
p(\mathbf{z}_t \mid W, A) = \frac{1}{Z(W,A)} \left( \prod_{i=1}^{N} A_{i, z_{i,t}} \right) \exp\left[ \frac{\beta}{2} \sum_{i=1}^{N} \sum_{j=1}^{N} W_{ij} \, \mathbb{I}(z_{i,t} = z_{j,t}) \right],$
where $\mathbf{z}_t = (z_{1,t}, \dots, z_{N,t})$ denotes the vector of latent-state assignments at step $t$, with $z_{i,t} \in \{1, \dots, K\}$ indicating the assignment for sample $x_i$. The hyperparameter $\beta$ controls the strength of graph propagation, and $Z(W,A)$ is the partition function.

\subsubsection{State-Level Success Model.} To statistically infer sample difficulty and latent-state assignments, we further integrate their relationship with historical rollout feedback into a Bayesian framework. However, estimating an independent success probability for each sample would still require a large number of sample-level rollouts to yield stable estimates. To address this and improve the efficiency of historical feedback utilization, we assume that samples assigned to the same latent state share a common state-level success probability.
For each latent state $k$, following~\citet{zeng2026cures} and~\citet{qu2026can}, we model its success probability at step $t$ with a Beta distribution:
$\theta_{k,t} \sim \operatorname{Beta}\bigl(a_{k,t}^{\mathrm{hist}}, b_{k,t}^{\mathrm{hist}}\bigr)$.
This distribution jointly captures the estimate of the state-level success probability while enabling sequential Bayesian updates.
We set the prior parameters for step $t$ by inheriting the posterior parameters from step $t-1$, i.e., $a_{k,t}^{\mathrm{hist}} = a_{k,t-1}$ and $b_{k,t}^{\mathrm{hist}} = b_{k,t-1}$.
Both parameters are initialized to $1$: $a_{k,0} = 1$, $b_{k,0} = 1$.
Conditioned on the latent-state assignment $z_{i,t}=k$ and the conditional independence of rollout outcomes, the number of successes $s_{i,t}$ out of $n_{i,t}$ rollouts follows a binomial distribution:
$(s_{i,t} \mid n_{i,t}, z_{i,t}=k, \theta_{k,t}) \sim \operatorname{Binomial}(n_{i,t}, \theta_{k,t})$.
Let $\boldsymbol{\theta}_t = (\theta_{1,t}, \dots, \theta_{K,t})$ collect the success probabilities of all latent states.
By combining the graph-structured prior over $\mathbf{z}_t$, the independent Beta priors over each $\theta_{k,t}$, and the sample-level binomial likelihoods, we obtain the joint distribution at step $t$:
$
p_t(\mathbf{s}_t, \mathbf{z}_t, \boldsymbol{\theta}_t \mid \mathbf{n}_t, W, A)
=\; p(\mathbf{z}_t \mid W, A)
\prod_{k=1}^{K} \operatorname{Beta}\bigl(\theta_{k,t} \mid a_{k,t}^{\mathrm{hist}}, b_{k,t}^{\mathrm{hist}}\bigr)  \prod_{i \in \mathcal{O}_t} \operatorname{Binomial}\bigl(s_{i,t} \mid n_{i,t}, \theta_{z_{i,t},t}\bigr)
$.

\subsubsection{Posterior Inference.} From this joint distribution, we apply Bayes' rule to infer the posterior over the latent-state assignments and state-level success probabilities after observing the rollout outcomes. This posterior allows us to jointly update our beliefs about each sample's latent state and the shared success probabilities, which is essential for subsequent difficulty estimation. It factorizes as $
p_t(\mathbf{z}_t, \boldsymbol{\theta}_t \mid \mathbf{s}_t, \mathbf{n}_t, W, A)
= \frac{p_t(\mathbf{s}_t, \mathbf{z}_t, \boldsymbol{\theta}_t \mid \mathbf{n}_t, W, A)}
{p_t(\mathbf{s}_t \mid \mathbf{n}_t, W, A)}
= \frac{p_t(\mathbf{s}_t, \mathbf{z}_t, \boldsymbol{\theta}_t \mid \mathbf{n}_t, W, A)}
{ \sum_{\mathbf{z}_t} \int p_t(\mathbf{s}_t, \mathbf{z}_t, \boldsymbol{\theta}_t \mid \mathbf{n}_t, W, A) \, \mathrm{d}\boldsymbol{\theta}_t}$.

\subsection{3.5 Online Mean-Field Variational Inference}

In this section, we employ online mean-field variational inference to support sample difficulty estimation for adaptive LLM RL scheduling. At each training step, only selected samples provide rollout feedback. The sample graph allows this feedback to update the latent-state distributions of related samples. Specifically, we approximate the joint posterior of the latent states $\mathbf{z}_t$ and state-level success probabilities $\boldsymbol{\theta}_t$. However, exact posterior inference is intractable. The Potts prior couples the latent states of neighboring samples. This coupling requires a summation over all $K^N$ possible assignments. Its computational cost grows exponentially with the number of training samples. To obtain tractable updates, we follow prior work~\cite{zhang1992meanmrf,hoffman2013stochastic} and adopt the mean-field approximation $q_t(\mathbf{z}_t,\boldsymbol{\theta}_t)\approx \prod_{i=1}^{N}q_i(z_{i,t})
\prod_{k=1}^{K}q(\theta_{k,t}).$Here, $q_i(z_{i,t})$ represents the latent-state distribution of sample $i$. We denote its state-assignment probability by $q_{ik,t}\equiv q_i(z_{i,t}=k).$ We restrict each state-level factor to the Beta family: $q(\theta_{k,t})=\operatorname{Beta}(a_{k,t},b_{k,t}).$

Let $\mathcal{Q}$ denote the variational family defined above. We seek the member of $\mathcal{Q}$ that minimizes the Kullback--Leibler (KL) divergence from the exact posterior:
$q_t^\star=\arg\min_{q_t\in\mathcal{Q}}\operatorname{KL}\left[q_t(\mathbf{z}_t,\boldsymbol{\theta}_t)\,\middle\|,p_t(\mathbf{z}_t,\boldsymbol{\theta}_t\mid\mathbf{s}_t,\mathbf{n}_t,W,A)\right].$
The exact posterior contains the partition function $Z(W,A)$ of the Potts prior. This makes the KL divergence difficult to evaluate directly. We instead use the evidence lower bound (ELBO):
$\mathcal{L}_t(q_t)=\mathbb{E}_{q_t}[\log p_t(\mathbf{s}_t,\mathbf{z}_t,\boldsymbol{\theta}_t\mid\mathbf{n}_t,W,A)]-\mathbb{E}_{q_t}[\log q_t(\mathbf{z}_t,\boldsymbol{\theta}_t)].$
The log marginal likelihood satisfies
$\log p_t(\mathbf{s}_t\mid\mathbf{n}_t,W,A)=\mathcal{L}_t(q_t)+\operatorname{KL}\left[q_t(\mathbf{z}_t,\boldsymbol{\theta}_t)\,\|,p_t(\mathbf{z}_t,\boldsymbol{\theta}_t\mid\mathbf{s}_t,\mathbf{n}_t,W,A)\right].$
The left-hand side does not depend on $q_t$. Thus, minimizing the KL divergence is equivalent to maximizing the ELBO:
$q_t^\star=\arg\max_{q_t\in\mathcal{Q}}\mathcal{L}_t(q_t).$

We can evaluate the ELBO under the mean-field approximation. In particular, the expected Potts interaction becomes
$\mathbb{E}_{q_t}\left[\mathbb{I}(z_{i,t}=z_{j,t})\right]=\sum_{k=1}^{K}q_{ik,t}q_{jk,t}.$
Substituting this result and the variational factors into the ELBO gives
\begin{align*}
\mathcal{L}_t(q)& = (\beta/2) \textstyle\sum_{i=1}^N \textstyle\sum_{j=1}^N W_{ij} \textstyle\sum_{k=1}^K q_{ik,t}q_{jk,t} \\
&+ \textstyle\sum_{i=1}^N \textstyle\sum_{k=1}^K q_{ik,t}\log A_{ik} - \textstyle\sum_{i=1}^N \textstyle\sum_{k=1}^K q_{ik,t}\log q_{ik,t} \\
&+ \textstyle\sum_{i\in\mathcal{O}_t} \textstyle\sum_{k=1}^K q_{ik,t} \bigl\{ s_{i,t}[\psi(a_{k,t})-\psi(a_{k,t}+b_{k,t})] \\
&\qquad + (n_{i,t}-s_{i,t})[\psi(b_{k,t})-\psi(a_{k,t}+b_{k,t})] \bigr\} \\
&- \textstyle\sum_{k=1}^K \operatorname{KL}[\operatorname{Beta}(a_{k,t},b_{k,t}) \| \operatorname{Beta}(a_{k,t}^{\mathrm{hist}},b_{k,t}^{\mathrm{hist}})].
\end{align*}
We omit terms independent of $q_t$. Here, $\psi(\cdot)$ denotes the digamma function. Appendix~A provides the detailed derivation.
The ELBO contains two groups of variational parameters. The first group contains the assignment probabilities $q_{ik,t}$. The second contains the Beta parameters $(a_{k,t},b_{k,t})$. We optimize one group while keeping the other fixed. This procedure is known as coordinate ascent. Proposition1 derives the assignment update, and Proposition~2 derives the Beta update. We alternate these updates during inference. Proposition~3 establishes the convergence of this alternating procedure. Appendix A provides the proofs.
\subsubsection{Proposition 1: Coordinate-Optimal Assignment Update}
Fix $q(\boldsymbol{\theta}_t)$ and all $q_j(z_{j,t})$ for $j\neq i$. The coordinate-optimal update is
$q_{ik,t}^{\star}=\operatorname{Softmax}_{k}[\log A_{ik}+\beta\sum_{j=1}^{N}W_{ij}q_{jk,t}+\mathbb{I}i\in\mathcal{O}_t)\ell_{ik,t}^{\mathrm{VB}}],$
where
$
\ell_{ik,t}^{\mathrm{VB}}
={}
s_{i,t}
\left[
\psi(a_{k,t})
-
\psi(a_{k,t}+b_{k,t})
\right]+
(n_{i,t}-s_{i,t})
\left[
\psi(b_{k,t})
-
\psi(a_{k,t}+b_{k,t})
\right].$
\subsubsection{Proposition 2: Coordinate-Optimal Beta Update}
Fix $q(\mathbf{z}_t)$. The coordinate-optimal factor is
$q^\star(\theta_{k,t})=\operatorname{Beta}(a_{k,t},b_{k,t}),$
where
$a_{k,t}=a_{k,t}^{\mathrm{hist}}+\sum_{i\in\mathcal{O}_t}q_{ik,t}s_{i,t},$
and
$b_{k,t}=b_{k,t}^{\mathrm{hist}}+\sum_{i\in\mathcal{O}_t}q_{ik,t}(n_{i,t}-s_{i,t}).$
\subsubsection{Proposition 3: Monotonicity and Convergence}
At training step $t$, we keep $(a_{k,t}^{\mathrm{hist}},b_{k,t}^{\mathrm{hist}})$ fixed. Each update in Propositions~1 and~2 maximizes the ELBO with respect to one group of parameters. Therefore, neither update decreases $\mathcal{L}_t(q)$ and the ELBO sequence converges. 
This convergence provides a stable estimator state at each LLM RL step (though it may not be globally optimal).

We now apply Propositions~1 and~2 in an online alternating procedure. After training step $t$ produces rollout feedback for $\mathcal{O}_t$, we initialize inference with the previous estimator state:
$q_{ik,t}^{(0)}=q_{ik,t-1}$, $
a_{k,t}^{(0)}=a_{k,t-1}$, $
b_{k,t}^{(0)}=b_{k,t-1}.$·
We keep $(a_{k,t}^{\mathrm{hist}},b_{k,t}^{\mathrm{hist}})$ fixed during inference at step $t$. 
Each iteration first applies Proposition~1 to update $q_{ik,t}$. It then applies Proposition~2 to update $(a_{k,t},b_{k,t})$. The first step updates the latent-state assignments and serves as an E-like update. The second updates the state-level parameters and serves as an M-like update. We repeat both steps until the ELBO increment falls below a preset threshold or the iteration count reaches its maximum.\footnote{We use a numerical approximation to accelerate computation. Code and Data Supplement provide the details.}
Proposition~1 updates the assignments of both observed and unobserved samples. Proposition~2 updates the state-level success probabilities using the observed feedback. Together, they transform the feedback from $\mathcal{O}_t$ into an updated estimator state for the full training set:
$\mathcal{S}_t
=
\{q_t,a_t,b_t\}.$
The next subsection converts $\mathcal{S}_t$ into sample-level difficulty estimates for adaptive scheduling. We also use $\mathcal{S}_t$ to initialize inference at the next training step.

\begin{table*}[h]
\centering
\begin{tabular}{ccccccccc}
\hline
\multirow{2}{*}{Method} & \multicolumn{4}{c}{Qwen-2.5-Math-1.5B (Average@8)} & \multicolumn{4}{c}{Llama-3.2-1B-Instruct (Average@8)} \\ \cline{2-9} 
 & MATH500 & AIME24 & AIME25 & Olym & MATH500 & AIME24 & AIME25 & Olym \\ \hline
Base & 39.3 & 3.30 & 3.33 & 25.2 & 17.5 & 0.00 & 0.00 & 3.02 \\
GVM & 71.9 & 11.3 & 11.3 & 35.8 & 23.9 & 0.42 & 0.00 & 4.59 \\
GVM+Ours& \underline{74.7} & \underline{16.7} & \underline{13.3} & \underline{38.3} & \underline{25.7} & \underline{3.30} & 0.00 & \underline{4.89} \\
PCL & 59.7 & 7.92 & 4.58 & 26.1 & 26.7 & \underline{0.42} & 0.00 & 5.04 \\
PCL+Ours & \underline{61.6} & \underline{11.3} & \underline{6.67} & \underline{27.1} & \underline{27.4} & 0.00 & \underline{0.42} & \underline{5.81} \\
GRESO & 66.8 & \underline{10.0} & 7.50 & 30.3 & 30.0 & 2.50 & 0.00 & 7.59 \\
GRESO+Ours & \underline{68.2} & \underline{10.0} & \underline{10.0} & \underline{31.9} & \underline{32.6} & \underline{6.67} & \underline{0.42} & \underline{8.26} \\ \hline
\end{tabular}
\caption{The Average@8 accuracy of different RL scheduling methods on four mathematical reasoning benchmarks using two base models. The first column lists the original schedulers and their variants integrated with our difficulty estimator. Underlined results indicate improvements over the corresponding original scheduler ($p < 5\times10^{-5}$, see Appendix C).}
\label{tab:overall}
\end{table*}

\subsection{3.6 Difficulty Estimation for Adaptive RL Schedulers}
In this section, we aggregate the state-level posterior from the previous training step into sample-level estimates of success probability, difficulty. These estimates then guide adaptive sample selection and rollout allocation. At the beginning of RL training step $t$, the estimator loads the state $\mathcal{S}_{t-1} = \{q_{t-1}, a_{t-1}, b_{t-1}\}$ from the previous step.
For each latent state $k$, we compute the posterior mean of its success probability as 
$
\mu_{k,t-1} = \frac{a_{k,t-1}}{a_{k,t-1}+b_{k,t-1}}
$
,
We first form a model-based prediction for sample $x_i$: $\widehat{p}_{i,t}^{\text{model}} = \sum_{k=1}^{K} q_{ik,t-1} \mu_{k,t-1}$.
Then we adjust the prediction based on whether the sample has been rolled out before.
If $x_i$ has no prior rollouts, we directly use the model prediction:
$\widehat{p}_{i,t} = \widehat{p}_{i,t}^{\text{model}}$.
If it has been sampled in earlier steps, we average the model prediction with its historical success rate $\bar r_{i,<t}$:
$\widehat{p}_{i,t} = \gamma \widehat{p}_{i,t}^{\text{model}} + (1-\gamma)\bar r_{i,<t} )$ ($\gamma =0.5$, see discussion in Appendix E).
Also, the estimated difficulty is then $\widehat{d}_{i,t} = 1 - \widehat{p}_{i,t}$.

Overall, before training step $t$, the estimator predicts the difficulty of every sample. These outputs serve as a plug-and-play component for most sample-level scheduling and resource-allocation frameworks in RLVR training.

\section{Experiment}
\subsection{Implementation Details}
\textbf{Setting.}
We conduct all training and inference on 4$\times$A100 80G GPUs.
To avoid data leakage~\cite{wu2026reasoning}, following ~\citet{yao2026optimizinggvm} and~\citet{zeng2026cures}, we adopt Qwen~2.5~1.5B~Math~\cite{yang2024qwen25mathtechnicalreportmathematical} and Llama~3~1B-Instruct~\cite{grattafiori2024llama3herdmodels} as base models. For the main experiments, we keep the environment strictly consistent with the original baselines.
We integrate our framework as a plug-and-play component into the original code to demonstrate that it improves the final RL performance. For fairness, we use the same computational budget for all experiments, equivalent to one round of the standard GVM algorithm ($\sim$1.5M rollouts, including pre-sampling).
We use Qwen3-0.6B~\cite{zhang2025qwen3embeddingadvancingtext} as the sentence embedding model, and set $K=320$, $k_{\mathrm{nn}}=80$, $\gamma=0.5$ and $\beta=2$.
Guidelines for hyperparameter or embedding model selection across different setting are provided in Appendix E.

\textbf{Dataset.}
Since all baselines we integrate are based on mathematical problems, we follow the experimental settings of previous works~\cite{yao2026optimizinggvm,pcl,zheng2026actgreso} and conduct reinforcement learning on NuminaMath~\cite{li2024numinamath,cui2025process}. We evaluate the resulting models on MATH500~\cite{hendrycks2021measuringmath,verifystepbystep}, AIME 2024~\cite{balunovic2026matharenaaime2024}, AIME 2025~\cite{dekoninck2026beyondaime2025}, and OlympiadBench~\cite{he2024olympiadbench}. We provide detailed descriptions of these datasets in Appendix B and discuss their generalizability to other domains in Appendix D.

\textbf{Baseline.}
We evaluate our plug-and-play component on three representative baselines from different scheduling paradigms: GVM, a rollout-allocation method~\cite{yao2026optimizinggvm}; PCL, a curriculum-learning method that trains a dedicated difficulty predictor~\cite{pcl}; and GRESO, a curriculum-learning method that uses historical success rates as difficulty signals~\cite{zheng2026actgreso}. For each baseline, we replace only its original difficulty estimator with ours.

\textbf{Metric.}
We use rule-based matching to judge whether the result matches the ground truth~\cite{guo2025deepseekr1,sheng2024hybridflow}. Because a single evaluation on small and challenging datasets such as AIME can be sensitive to sampling randomness, we follow prior work~\cite{yao2026optimizinggvm,zheng2026actgreso} and sample eight independent rollouts per prompt, reporting Average@8 accuracy.

\subsection{Overall Performance}
We integrate our estimator into three representative RL schedulers: GVM for rollout allocation, PCL for curriculum selection, and GRESO for selective rollout. We replace their original difficulty estimators while retaining their scheduling strategies, and compare all methods under the same total generation budget.

As shown in Table~\ref{tab:overall}, our estimator improves the overall performance of all three schedulers, with gains in most model--dataset settings. For GVM, GVM+Ours retains pre-sampling only for gradient-related estimation and replaces the original difficulty estimate with ours. It improves performance in seven of the eight settings and matches the baseline in the remaining one. We will provide a more in-depth analysis in the next section. When integrated into PCL and GRESO, our estimator also improves the corresponding baselines in most settings.
A plausible explanation is that the more stable difficulty estimates help the schedulers identify samples that are likely to yield both correct and incorrect responses within the same rollout group. Such samples produce nondegenerate group-relative advantages and potentially more informative optimization signals.

Although full-history aggregation may introduce temporal lag, graph propagation and shared state-level statistics make its effect largely systematic across related samples, affecting absolute calibration more than relative difficulty ranking. Since adaptive schedulers primarily rely on ranking for sample selection and rollout allocation, this bias has limited impact on scheduling, as supported by the high correlations and downstream gains above. Appendix~G further examines a windowed variant that reduces stale historical influence.

Overall, these results suggest that our estimator can serve as a reusable component for different RL scheduling strategies and improve downstream performance under a matched generation budget.

\subsection{Can Our Method Continuously Track Sample Difficulty During RL Training?}

\begin{table*}[h]
\small
\centering
\begin{tabular}{ccccccccccc}
\hline
\multirow{2}{*}{Base Model} & \multirow{2}{*}{Methods} & \multicolumn{2}{c}{Early Steps} & \multicolumn{2}{c}{Middle Steps} & \multicolumn{2}{c}{Last Steps} & \multicolumn{2}{c}{Full Steps} & \multirow{2}{*}{Time (A100*h)} \\ \cline{3-10}
 &  & MAE ($\downarrow$) & r ($\uparrow$) & MAE ($\downarrow$) & r ($\uparrow$) & MAE ($\downarrow$) & r ($\uparrow$) & MAE ($\downarrow$) & r ($\uparrow$) &  \\ \hline
\multirow{5}{*}{\begin{tabular}[c]{@{}c@{}}Qwen-2.5\\ Math-1.5B\end{tabular}} & Sample before RL & 0.116 & 0.544 & 0.139 & 0.856 & 0.151 & 0.865 & 0.135 & 0.522 & $\sim$29.6 \\
 & Sample before Step & \textbf{0.063} & \textbf{0.959} & \textbf{0.054} & \textbf{0.971} & \textbf{0.049} & \textbf{0.974} & \textbf{0.055} & \textbf{0.969} & $\sim$45.3 \\
 & VIP & 0.373 & - & \underline{0.094} & 0.794 & \underline{0.084} & 0.885 & 0.184 & 0.423 & $\sim$0.05 \\
 & PCL & 0.404 & 0.138 & 0.336 & 0.407 & 0.328 & 0.474 & 0.356 & 0.281 & $\sim$7.36 \\
  & MoPPS & 0.373 & - & 0.149 & 0.910 & 0.118 & 0.929 & 0.214 & 0.754 & - \\
 & Ours & \underline{0.290} & \underline{0.482} & 0.133 & \underline{0.912} & 0.123 & \underline{0.943} & \underline{0.183} & \underline{0.836} & $\sim$0.12 \\ \hline
\multirow{5}{*}{\begin{tabular}[c]{@{}c@{}}Llama-3.2\\ 1B-Instruct\end{tabular}} & Sample before RL & 0.116 & 0.374 & 0.141 & 0.648 & 0.154 & 0.671 & 0.137 & 0.394 & $\sim$33.4 \\
 & Sample before Step & \textbf{0.044} & \textbf{0.986} & \textbf{0.046} & \textbf{0.967} & \textbf{0.046} & \textbf{0.967} & \textbf{0.045} & \textbf{0.986} & $\sim$47.9 \\
 & VIP & 0.413 & - & 0.107 &  \underline{0.895} & 0.084 & 0.880 & 0.201 & 0.717 & $\sim$0.05 \\
& PCL & 0.204 & 0.467 & 0.181 & 0.404 & 0.187 & 0.525 & 0.191 & -0.231 & $\sim$4.38 \\
 & MoPPS & 0.413 & - & 0.133 & 0.864 & 0.109 & 0.869 & 0.218 & 0.606 & - \\
 & Ours & \underline{0.197} & \underline{0.469} & \underline{0.100} & 0.868 & \underline{0.075} & \underline{0.913} & \underline{0.131} & \underline{0.776} & $\sim$0.12 \\ \hline
\end{tabular}

\caption{Difficulty-estimation accuracy and overhead at different stages of GRPO training (batch size =256). MAE is computed at the sample level and Pearson correlation $r$ at the batch level. Bold denotes the best overall result, underlining denotes the best result among low-cost methods, and Time (including embedding, clustering and inference) denotes additional estimation overhead for over 3 epochs on 150K training samples.}
\label{tab:analysis}
\end{table*}

We evaluate how accurately different methods track evolving sample difficulty along the same GRPO training trajectory. At each evaluation point, we use the empirical success rate from 8 independent rollouts generated by the corresponding policy as the reference empirical success probability. We report sample-level MAE for calibration and batch-level Pearson correlation $r$ (batch size 256) for ranking consistency in Table~\ref{tab:analysis}, since sample-level empirical rates are discrete and sparse. Lower MAE and higher $r$ indicate better estimation.

We compare two high-cost probing baselines, Sample Before RL (SBR) and Sample Before Step (SBS), with three representative low-cost approaches: PCL (LLM-based difficulty prediction), VIP (historical-feedback-based statistical estimation)
, and MoPPS (history-based dynamic modeling).
SBS probes samples repeatedly with the current policy; it therefore achieves the highest accuracy but requires roughly 45--48 hours of extra computation.
SBR, however, estimates difficulty only once before training, so its predictions quickly become outdated as the policy evolves, highlighting the limitation of static estimates.
Our approach performs best overall among low-cost methods, with a particularly clear advantage in the early sparse-feedback stage. Early in training, it achieves Pearson correlations of $0.482$ for Qwen and $0.469$ for Llama, together with the lowest MAE among the low-cost baselines. These results indicate that the graph structure alleviates observation sparsity by transferring feedback across related samples. As more rollout observations accumulate, the full-trajectory correlations increase to $0.836$ and $0.776$, while the total estimation overhead remains approximately 0.12h.
Overall, our method mitigates early-stage observation sparsity and tracks policy-induced difficulty changes with low computational overhead.
\subsection{Ablation Study}
\begin{table}[]
\small
\centering
\begin{tabular}{cccc}
\hline
\multirow{2}{*}{Base Model} & \multirow{2}{*}{Methods} & \multicolumn{2}{c}{Full Steps} \\
 &  & MAE (↓) & r (↑) \\ \hline
\multirow{4}{*}{\begin{tabular}[c]{@{}c@{}}Qwen-2.5\\ Math-1.5B\end{tabular}} & Ours & \textbf{0.183} & \textbf{0.836} \\
 & w/o Spectral Init. & 0.218 & 0.710 \\
 & w/o Prior Smoothing & 0.197 & 0.783 \\
 & w/o Graph Sparsification & 0.187 & 0.835 \\ \hline
\multirow{4}{*}{\begin{tabular}[c]{@{}c@{}}Llama-3.2\\ 1B-Instruct\end{tabular}} & Ours & \textbf{0.131} & \textbf{0.776} \\
 & w/o Spectral Init. & 0.197 & 0.686 \\
 & w/o Prior Smoothing & 0.144 & 0.736 \\
 & w/o Graph Sparsification & 0.134 & 0.742 \\ \hline
\end{tabular}
\caption{Full-trajectory ablation results for difficulty estimation on two base models. Lower MAE and higher Pearson correlation $r$ indicate better estimation.}
\label{tab:ablation}
\end{table}

We ablate spectral initialization, prior smoothing, and graph sparsification. Table~\ref{tab:ablation} reports results over the full training trajectory.
Random initialization causes a substantial performance drop. Since our EM-style optimization only guarantees convergence to a local optimum, spectral clustering provides a high-quality starting point that guides inference toward a better solution, whereas random initialization can lead to poor local optima.
Replacing the smoothed latent-state prior $A_{ik}$ with hard assignments also degrades performance. As the policy evolves, sample difficulty and its latent-state assignment should evolve accordingly. The smoothed prior preserves the initial clustering structure while allowing rollout feedback to revise the assignments, whereas hard assignments prevent such adaptation.
Using a dense graph causes a modest decline because weakly related edges introduce noise into feedback propagation. Graph sparsification retains more reliable relationships between samples.

\subsection{Do Spectral Clusters Capture Semantic and Difficulty Structure?}

We use MATH~\cite{hendrycks2021measuringmath} for this analysis because it provides standardized annotations of both subject category and difficulty level, allowing us to evaluate the semantic and difficulty information captured by the clusters separately. Following Sec.3.3, we embed the MATH problems and perform spectral clustering. The resulting clusters are strongly associated with subject categories ($\chi^2=10{,}995.84$, $p<10^{-300}$; Cramér's $V=0.494$), indicating the spectral partition effectively captures semantic structure. After controlling for subject category, difficulty distributions still differ significantly across clusters (Kruskal--Wallis, $p_{max}<5\times10^{-3}$). These results show the clusters capture not only problem semantics but also difficulty information beyond subject identity, thereby providing an informative initial structure for subsequent online difficulty inference. We provide the full statistical details in Appendix F.

\section{Conclusion}

In this work, we address inefficient exploration-budget allocation in RLVR through graph-based online difficulty estimation. Our framework connects samples with similar semantic content and reasoning structures and uses rollout feedback to track their evolving difficulty without dedicated probing. The resulting estimates support both sample-selection and rollout-allocation schedulers. Experiments across multiple base models, schedulers, and benchmarks show that our framework improves downstream reasoning performance in most settings under matched rollout budgets while adding little online computational overhead. These results demonstrate the potential of online difficulty estimation to support more efficient RLVR training.

\bigskip

\section{A. ELBO Derivation and Proofs}

This section derives the expanded evidence lower bound (ELBO) in Sec.~3.5 and proves Propositions~1--3. For compactness, define
$\delta_{ik,t}=\mathbb{I}(z_{i,t}=k)$ and let $C_t$ collect all terms that are independent of the variational distribution. The mutual-neighbor graph is undirected, so $W_{ij}=W_{ji}$, and it has no self-loops, so $W_{ii}=0$.

\subsection{A.1 Derivation of the Expanded ELBO}

From the joint model in Sec.~3.4, its log density can be written, up to the constant $C_t$, as
{\small
\begin{align}
&\log p_t(\mathbf{s}_t,\mathbf{z}_t,\boldsymbol{\theta}_t
\mid\mathbf{n}_t,W,A) \nonumber\\
={}& C_t
+\sum_{i=1}^{N}\sum_{k=1}^{K}\delta_{ik,t}\log A_{ik}
+\frac{\beta}{2}\sum_{i=1}^{N}\sum_{j=1}^{N}W_{ij}
  \sum_{k=1}^{K}\delta_{ik,t}\delta_{jk,t} \nonumber\\
&+\sum_{k=1}^{K}\left[
(a_{k,t}^{\mathrm{hist}}-1)\log\theta_{k,t}
+(b_{k,t}^{\mathrm{hist}}-1)\log(1-\theta_{k,t})
\right] \nonumber\\
&+\sum_{i\in\mathcal{O}_t}\sum_{k=1}^{K}
\delta_{ik,t}s_{i,t}\log\theta_{k,t} \nonumber\\
&+\sum_{i\in\mathcal{O}_t}\sum_{k=1}^{K}\delta_{ik,t}
(n_{i,t}-s_{i,t})\nonumber\\[-2pt]
&\qquad\qquad\qquad{}\times\log(1-\theta_{k,t}).
\label{eq:appendix-log-joint}
\end{align}
}
Here, $C_t$ includes $-\log Z(W,A)$, the Beta normalizers, and the binomial coefficients, all of which are fixed with respect to $q_t$.

Under the mean-field distribution,
\begin{align}
\mathbb{E}_{q_t}[\delta_{ik,t}]&=q_{ik,t},\\
\mathbb{E}_{q_t}[\delta_{ik,t}\delta_{jk,t}]&=q_{ik,t}q_{jk,t}
\quad (i\neq j),\\
\mathbb{E}_{q_t}[\log\theta_{k,t}]&=
\psi(a_{k,t})-\psi(a_{k,t}+b_{k,t}),\\
\mathbb{E}_{q_t}[\log(1-\theta_{k,t})]&=
\psi(b_{k,t})-\psi(a_{k,t}+b_{k,t}).
\end{align}
The second identity is sufficient for the graph term because $W_{ii}=0$. The negative entropy contribution factorizes as
\begin{align}
-\mathbb{E}_{q_t}[\log q_t]
={}&-\sum_{i=1}^{N}\sum_{k=1}^{K}q_{ik,t}\log q_{ik,t}\nonumber\\
&-\sum_{k=1}^{K}\mathbb{E}_{q_t}[\log q(\theta_{k,t})].
\end{align}
Combining the expected log Beta prior with the corresponding Beta-factor entropy gives
\begin{align}
&\mathbb{E}_{q_t}\!\left[
\log \operatorname{Beta}(\theta_{k,t}\mid
a_{k,t}^{\mathrm{hist}},b_{k,t}^{\mathrm{hist}})
-\log q(\theta_{k,t})\right] \nonumber\\
&\qquad=-\operatorname{KL}\!\left[
\operatorname{Beta}(a_{k,t},b_{k,t})
\,\middle\|\,
\operatorname{Beta}(a_{k,t}^{\mathrm{hist}},b_{k,t}^{\mathrm{hist}})
\right].
\end{align}
Substituting these identities into
$\mathcal{L}_t(q_t)=\mathbb{E}_{q_t}[\log p_t]-\mathbb{E}_{q_t}[\log q_t]$
yields the expression below. To keep the display compact, let
\begin{align}
D_{k,t}^{\mathrm{Beta}}
=\operatorname{KL}\!\left[
\operatorname{Beta}(a_{k,t},b_{k,t})
\,\middle\|\,
\operatorname{Beta}(a_{k,t}^{\mathrm{hist}},b_{k,t}^{\mathrm{hist}})
\right].
\end{align}
Then
\begin{align}
\mathcal{L}_t(q_t)
={}&\frac{\beta}{2}\sum_{i=1}^{N}\sum_{j=1}^{N}W_{ij}
\sum_{k=1}^{K}q_{ik,t}q_{jk,t}\nonumber\\
&+\sum_{i=1}^{N}\sum_{k=1}^{K}q_{ik,t}\log A_{ik}\nonumber\\
&-\sum_{i=1}^{N}\sum_{k=1}^{K}q_{ik,t}\log q_{ik,t}\nonumber\\
&+\sum_{i\in\mathcal{O}_t}\sum_{k=1}^{K}
q_{ik,t}\ell_{ik,t}^{\mathrm{VB}}\nonumber\\
&-\sum_{k=1}^{K}D_{k,t}^{\mathrm{Beta}}+C_t,
\label{eq:appendix-elbo}
\end{align}
where
\begin{align}
\ell_{ik,t}^{\mathrm{VB}}
={}&s_{i,t}\left[\psi(a_{k,t})-\psi(a_{k,t}+b_{k,t})\right]\nonumber\\
&+(n_{i,t}-s_{i,t})
\left[\psi(b_{k,t})-\psi(a_{k,t}+b_{k,t})\right].
\end{align}
Dropping $C_t$, which is independent of $q_t$, gives exactly the expanded ELBO reported in Sec.~3.5.

\subsection{A.2 Proof of Proposition 1}

Fix $q(\boldsymbol{\theta}_t)$ and every assignment factor except $q_i(z_{i,t})$. Because $W$ is symmetric, the two appearances of node $i$ in the double graph sum combine and cancel the factor $1/2$. The terms of Eq.~\eqref{eq:appendix-elbo} that depend on $q_{ik,t}$ are therefore
\begin{align}
\mathcal{L}_t(q_i)
={}&\sum_{k=1}^{K}q_{ik,t}h_{ik,t}
-\sum_{k=1}^{K}q_{ik,t}\log q_{ik,t}+C,\\
h_{ik,t}
={}&\log A_{ik}
+\beta\sum_{j=1}^{N}W_{ij}q_{jk,t}
+\mathbb{I}(i\in\mathcal{O}_t)\ell_{ik,t}^{\mathrm{VB}}.
\end{align}
Introduce a Lagrange multiplier $\lambda_i$ for
$\sum_{k=1}^{K}q_{ik,t}=1$ and define
\begin{align}
\widetilde{\mathcal{L}}_t(q_i)
=\mathcal{L}_t(q_i)+\lambda_i
\left(\sum_{r=1}^{K}q_{ir,t}-1\right).
\end{align}
Differentiating gives
\begin{align}
\frac{\partial\widetilde{\mathcal{L}}_t(q_i)}
{\partial q_{ik,t}}
=h_{ik,t}-\log q_{ik,t}-1+\lambda_i.
\end{align}
Setting this derivative to zero and normalizing over $k$ yields
\begin{align}
q_{ik,t}^{\star}
=\frac{\exp(h_{ik,t})}{\sum_{r=1}^{K}\exp(h_{ir,t})}
=\operatorname{Softmax}_{k}(h_{ik,t}),
\end{align}
which is the update in Proposition~1.

\subsection{A.3 Proof of Proposition 2}

Fix $q(\mathbf{z}_t)$ and define
$q_{-k,t}=q(\mathbf{z}_t)\prod_{r\neq k}q(\theta_{r,t})$.
The standard mean-field coordinate identity gives
\begin{align}
\log q^{\star}(\theta_{k,t})
={}&\mathbb{E}_{q_{-k,t}}\bigl[
\log p_t(\mathbf{s}_t,\mathbf{z}_t,\boldsymbol{\theta}_t
\nonumber\\[-2pt]
&\qquad\mid\mathbf{n}_t,W,A)\bigr]+C.
\end{align}
Retaining only terms involving $\theta_{k,t}$ in Eq.~\eqref{eq:appendix-log-joint} gives
\begin{align}
\alpha_{k,t}&=a_{k,t}^{\mathrm{hist}}
+\sum_{i\in\mathcal{O}_t}q_{ik,t}s_{i,t},\\
\gamma_{k,t}&=b_{k,t}^{\mathrm{hist}}
+\sum_{i\in\mathcal{O}_t}q_{ik,t}(n_{i,t}-s_{i,t}),\\
\log q^{\star}(\theta_{k,t})
={}&(\alpha_{k,t}-1)\log\theta_{k,t}\nonumber\\
&+(\gamma_{k,t}-1)\log(1-\theta_{k,t})+C.
\end{align}
This is the log density of a Beta distribution. Hence
\begin{align}
q^{\star}(\theta_{k,t})&=\operatorname{Beta}(a_{k,t},b_{k,t}),\\
a_{k,t}&=a_{k,t}^{\mathrm{hist}}
+\sum_{i\in\mathcal{O}_t}q_{ik,t}s_{i,t},\\
b_{k,t}&=b_{k,t}^{\mathrm{hist}}
+\sum_{i\in\mathcal{O}_t}q_{ik,t}(n_{i,t}-s_{i,t}),
\end{align}
which proves Proposition~2.

\subsection{A.4 Sequential and Synchronous Implementations}

Proposition~1 gives the exact coordinate-optimal update for a
single assignment factor $q_i(z_{i,t})$ while all other factors are
held fixed. We consider two implementations of this update.

\paragraph{Sequential reference implementation.}
Our reference implementation performs a strict sequential
Gauss--Seidel sweep over the samples. Under the update order
$i=1,\ldots,N$, the assignment update at iteration $m+1$ is
\begin{align}
q_{ik,t}^{(m+1)}
=
\operatorname{Softmax}_{k}
\Bigg[
&\log A_{ik}
+\beta\sum_{j<i}W_{ij}q_{jk,t}^{(m+1)}
+\beta\sum_{j>i}W_{ij}q_{jk,t}^{(m)}
\nonumber\\
&+\mathbb{I}(i\in\mathcal{O}_t)
\ell_{ik,t}^{\mathrm{VB},(m)}
\Bigg].
\label{eq:sequential-assignment-update}
\end{align}
Thus, each update uses the most recently available values of the
previously updated factors. After one complete assignment sweep,
the Beta factors are updated according to Proposition~2.
Because every assignment factor is optimized while the remaining
factors are fixed, this implementation is an exact coordinate-ascent
variational inference procedure. The monotonicity and convergence
results in Proposition~3 apply to this sequential implementation.

\paragraph{Synchronous accelerated implementation.}
The strict sequential update is difficult to parallelize over
samples. We therefore additionally implement a synchronous
Jacobi-style update:
\begin{align}
\widetilde q_{ik,t}^{(m+1)}
=
\operatorname{Softmax}_{k}
\left[
\log A_{ik}
+\beta\sum_{j=1}^{N}W_{ij}q_{jk,t}^{(m)}
+\mathbb{I}(i\in\mathcal{O}_t)
\ell_{ik,t}^{\mathrm{VB},(m)}
\right].
\label{eq:synchronous-assignment-update}
\end{align}
All assignment factors in Eq.~\eqref{eq:synchronous-assignment-update}
are computed in parallel from the same previous iterate, after which
the Beta factors are updated using Proposition~2.

Unlike the sequential implementation, a synchronous assignment
sweep is not an exact block-coordinate maximization of the ELBO.
Consequently, Proposition~3 does not directly guarantee that every
synchronous sweep is ELBO non-decreasing. We treat the synchronous
implementation as a parallel approximation to the theoretically
grounded sequential procedure.

To evaluate this approximation, we compare the converged
sample-level success-probability estimates produced by the two
implementations. We define their discrepancy at training step $t$ as
\begin{align}
\Delta_t
=
\max_{1\leq i\leq N}
\left|
\widehat p_{i,t}^{\mathrm{sync}}
-
\widehat p_{i,t}^{\mathrm{seq}}
\right|.
\label{eq:sync-seq-discrepancy}
\end{align}
Across the evaluated training checkpoints, the discrepancy remains
within $10^{-2}$. At the same time, the synchronous implementation
allows the graph-message aggregation and assignment updates to be
executed efficiently in parallel, resulting in substantially better
wall-clock performance. We therefore use and recommend the
synchronous implementation in practice, while retaining the
sequential implementation as the reference algorithm for the
theoretical analysis.

\subsection{A.5 Proof of Proposition 3}

We prove the result for the strict sequential implementation
described in Sec.~A.4. Fix a training step $t$. Throughout the
inference procedure at this step, the historical parameters
$(a_{k,t}^{\mathrm{hist}},b_{k,t}^{\mathrm{hist}})$, the graph
$(W,A)$, and all rollout observations are fixed.

We assume that
\begin{align}
&A_{ik}>0
\quad\text{for every } i,k, \label{eq:assumption-positive-A}\\
&a_{k,t}^{\mathrm{hist}}>0,
\qquad
b_{k,t}^{\mathrm{hist}}>0
\quad\text{for every } k, \label{eq:assumption-positive-beta}\\
&|W_{ij}|<\infty,
\qquad
n_{i,t}<\infty. \label{eq:assumption-finite}
\end{align}
The first condition is satisfied by the smoothed initialization
when $0<\epsilon<1$. We further assume that the categorical factors
are updated cyclically and that every factor is updated once in each
complete sequential sweep.

\paragraph{Monotonicity.}
Fix all variational factors except one categorical factor
$q_i(z_{i,t})$. Proposition~1 gives the unique maximizer of the ELBO
with respect to this factor. Therefore, updating $q_i(z_{i,t})$
cannot decrease the ELBO.

Similarly, after fixing $q(\mathbf z_t)$ and all Beta factors except
$q(\theta_{k,t})$, Proposition~2 gives the coordinate-optimal Beta
factor. Updating this factor therefore cannot decrease the ELBO.
Consequently, if $q_t^{(m)}$ denotes the variational distribution
after the $m$-th coordinate update, then
\begin{align}
\mathcal{L}_t(q_t^{(m+1)})
\geq
\mathcal{L}_t(q_t^{(m)}).
\label{eq:elbo-monotonicity}
\end{align}

\paragraph{Boundedness of the parameter sequence.}
Each categorical parameter belongs to the probability simplex:
\begin{align}
(q_{i1,t},\ldots,q_{iK,t})\in\Delta_K.
\end{align}
The product of the $N$ categorical simplexes is compact.

Define the total number of observed successes and failures at
training step $t$ as
\begin{align}
S_t
&=
\sum_{i\in\mathcal O_t}s_{i,t},\\
F_t
&=
\sum_{i\in\mathcal O_t}(n_{i,t}-s_{i,t}).
\end{align}
By Proposition~2 and $0\leq q_{ik,t}\leq1$, the Beta parameters
satisfy
\begin{align}
a_{k,t}^{\mathrm{hist}}
\leq a_{k,t}
\leq a_{k,t}^{\mathrm{hist}}+S_t,\\
b_{k,t}^{\mathrm{hist}}
\leq b_{k,t}
\leq b_{k,t}^{\mathrm{hist}}+F_t.
\end{align}
Therefore, all Beta parameters remain in compact positive intervals.
The complete variational-parameter sequence consequently lies in
the compact set
\begin{align}
\mathcal X_t
={}&
\left(\prod_{i=1}^{N}\Delta_K\right)
\nonumber\\
&\times
\prod_{k=1}^{K}
\left[
a_{k,t}^{\mathrm{hist}},
a_{k,t}^{\mathrm{hist}}+S_t
\right]
\nonumber\\
&\times
\prod_{k=1}^{K}
\left[
b_{k,t}^{\mathrm{hist}},
b_{k,t}^{\mathrm{hist}}+F_t
\right].
\label{eq:compact-parameter-domain}
\end{align}
In particular, the parameter sequence has at least one accumulation
point.

\paragraph{Convergence of the ELBO values.}
Let
\begin{align}
p_t^{\mathrm{post}}
=
p_t(
\mathbf z_t,\boldsymbol\theta_t
\mid
\mathbf s_t,\mathbf n_t,W,A)
\end{align}
denote the exact posterior. The ELBO satisfies
\begin{align}
\mathcal L_t(q_t)
={}&
\log p_t(
\mathbf s_t
\mid
\mathbf n_t,W,A)
\nonumber\\
&-
\operatorname{KL}
\left[
q_t
\,\middle\|\,
p_t^{\mathrm{post}}
\right].
\label{eq:elbo-kl-decomposition}
\end{align}
Because the model has a finite number of latent assignments, proper
Beta priors, and finite rollout counts, the log evidence is finite.
Since the KL divergence is nonnegative,
\begin{align}
\mathcal L_t(q_t)
\leq
\log p_t(
\mathbf s_t
\mid
\mathbf n_t,W,A).
\end{align}
Combining this upper bound with
Eq.~\eqref{eq:elbo-monotonicity} shows that the ELBO-value sequence
is monotone and bounded above. Hence there exists a finite
$\mathcal L_t^\star$ such that
\begin{align}
\lim_{m\rightarrow\infty}
\mathcal L_t(q_t^{(m)})
=
\mathcal L_t^\star.
\label{eq:elbo-value-convergence}
\end{align}

\paragraph{Stationarity of accumulation points.}
Consider the iterates at the boundaries of complete sequential
sweeps, and let $T_t$ denote the mapping corresponding to one full
sweep of all categorical and Beta-factor updates. Under
Eqs.~\eqref{eq:assumption-positive-A}--\eqref{eq:assumption-finite},
the softmax assignment update, the Beta update, and the associated
digamma expectations are continuous on $\mathcal X_t$. Therefore,
$T_t$ is continuous.

Let $\bar q_t$ be an accumulation point of the sweep-level
variational sequence. Suppose, for contradiction, that
$T_t(\bar q_t)\neq\bar q_t$. Because every coordinate update is the
unique maximizer of its coordinate subproblem, at least one update
in the sweep would produce a strict ELBO increase. Hence
\begin{align}
\mathcal L_t(T_t(\bar q_t))
>
\mathcal L_t(\bar q_t).
\label{eq:strict-improvement-at-nonfixed-point}
\end{align}
By continuity, the same positive improvement would hold in a
neighborhood of $\bar q_t$. This contradicts the convergence of the
ELBO values in Eq.~\eqref{eq:elbo-value-convergence}, according to
which the ELBO improvement over a complete sweep must approach zero.
Therefore,
\begin{align}
T_t(\bar q_t)=\bar q_t.
\end{align}

Thus, every accumulation point is a fixed point of all exact
coordinate updates and is consequently coordinate-wise optimal.
Since the categorical updates are strictly positive under
$A_{ik}>0$, and the Beta parameters remain strictly positive, the
ELBO is differentiable at such a fixed point subject to the simplex
constraints. The corresponding first-order Karush--Kuhn--Tucker
conditions are therefore satisfied. Every accumulation point is
hence a stationary point of the ELBO.

The argument establishes convergence of the ELBO values and
stationarity of every accumulation point. 

\subsection{A.6 EM Iteration Visualization }

\begin{figure}
    \centering
    \includegraphics[width=\linewidth]{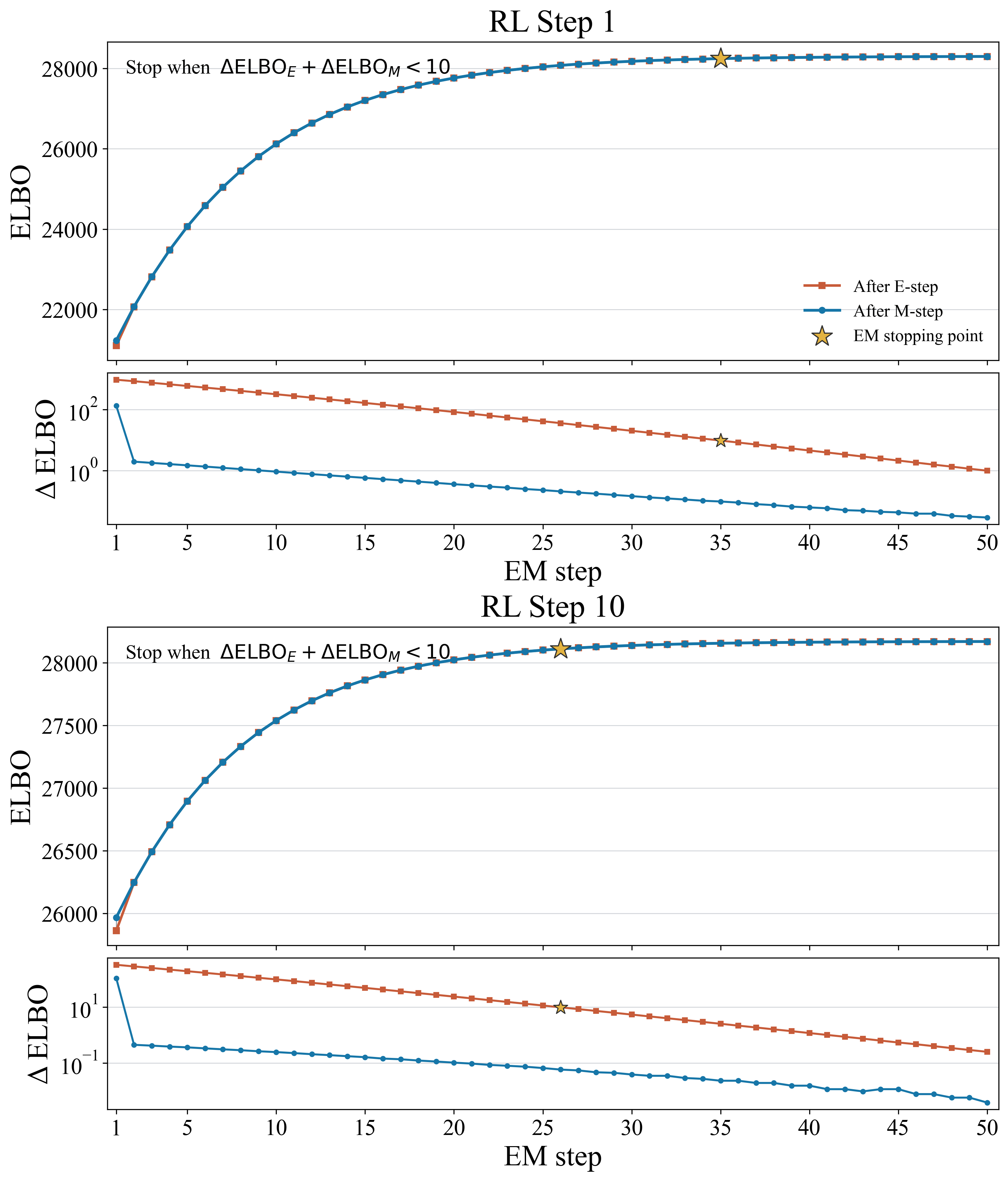}
    \caption{EM iterations}
    \label{fig:EM}
\end{figure}
We show the evolution of the ELBO under EM iterations in Figure~\ref{fig:EM}. In general, the M‑step performs a notably large update only at the very first iteration. The per‑round improvement in the ELBO from the E‑step and M‑step decays roughly exponentially, and convergence to a local stationary point or local maximum is typically reached within 50 iterations.
\section{B. Details of Datasets and Training}

\subsubsection{NuminaMath.} NuminaMath is a large-scale mathematical reasoning dataset comprising approximately 860K problem–solution pairs. It covers problems ranging from Chinese high-school mathematics to U.S. and international mathematical olympiad competitions, with data primarily collected from online examination papers and mathematics discussion forums. Each problem is accompanied by a chain-of-thought solution and a verifiable final answer. Following the data-processing protocols of GVM and CurES, we extract approximately 150K problems from the original corpus for RL training~\cite{cui2025process}. 

\subsubsection{MATH.} MATH is a competition-level mathematical reasoning dataset containing 12.5K problems, each accompanied by a complete step-by-step solution and a final answer. The dataset provides predefined difficulty annotations ranging from Level 1 to Level 5 and category labels covering seven subjects: prealgebra, algebra, number theory, counting and probability, geometry, intermediate algebra, and precalculus. We use these annotations only as external references rather than as training supervision. Specifically, we compare the complete method with variants that remove the clustering or graph-based components and examine how well the learned representations and structural relationships align with the provided category and difficulty labels. This analysis assesses whether the proposed components capture semantic similarity among problems and distinguish problems at different difficulty levels. 

MATH-500. MATH-500 is a representative subset of 500 problems selected from the original MATH benchmark. It retains the diverse mathematical subjects and difficulty levels of MATH while enabling more efficient and standardized evaluation. Each problem is accompanied by a reference solution and a verifiable final answer. We use MATH-500 to evaluate the models’ general mathematical reasoning ability across a broad range of competition-level problems. 

\subsubsection{AIME 2024.} Our AIME 2024 evaluation set combines the 15 problems from AIME I and the 15 problems from AIME II, yielding 30 problems in total. These problems cover areas such as algebra, geometry, number theory, counting, and probability and generally require multi-step reasoning and nontrivial mathematical insight. Each problem has a unique integer answer between 000 and 999, allowing reliable rule-based evaluation. We use AIME 2024 to assess reasoning performance on challenging competition-level mathematics problems. The general AIME format consists of 15 questions per examination, each requiring an integer answer from 000 to 999. 

\subsubsection{AIME 2025.} Our AIME 2025 evaluation set similarly combines AIME I and AIME II, resulting in 30 problems. Each problem requires an integer answer between 000 and 999 and involves advanced high-school mathematics across multiple domains. Since these problems were released more recently than most commonly used mathematical reasoning benchmarks, AIME 2025 provides a challenging evaluation of the models’ reasoning and generalization capabilities.

\subsubsection{OlympiadBench.} OlympiadBench is a bilingual and multimodal scientific reasoning benchmark containing 8,476 problems collected from international olympiads, Chinese olympiads, and the Chinese college entrance examination. The final ACL 2024 version contains 6,142 mathematics problems and 2,334 physics problems, with every problem accompanied by an expert-annotated solution. The benchmark includes both open-ended problems and theorem-proving problems and covers English and Chinese questions with textual or visual information. Following the evaluation protocol of GVM and CurES, we use the corresponding mathematics portion of OlympiadBench to evaluate reasoning performance on challenging olympiad-level problems.

\section{C. Significance Analysis of Main Results}

\subsection{Question and Comparison Units}

The main experiments report one Average@8 point estimate for each
scheduler--base-model--benchmark combination. We use a two-sided exact sign
test~\cite{dixon1946sign} to examine whether the observed improvements are
directionally consistent across these heterogeneous experimental settings.
This analysis directly evaluates whether integrating our method produces
positive changes more frequently than would be expected under an equal-probability
null hypothesis.

The analysis includes the three full integrations: GVM + Ours, PCL + Ours, and GRESO + Ours.

For integration method \(m\), base model \(b\), and benchmark \(d\), define
\begin{align}
\Delta_{m,b,d}
=
\operatorname{Acc}_{m+\mathrm{Ours},b,d}
-
\operatorname{Acc}_{m,b,d}.
\end{align}
We record a gain if \(\Delta_{m,b,d}>0\), a loss if
\(\Delta_{m,b,d}<0\), and a tie if \(\Delta_{m,b,d}=0\). Let \(W\), \(L\),
and \(T\) denote the resulting numbers of gains, losses, and ties,
respectively. Because ties provide no directional information, they are
excluded from the sign-test sample size, yielding \(n=W+L\).

\subsection{Exact Test and Results}

Under the null hypothesis that a non-tied comparison is equally likely to
favor either method,
\begin{align}
H_0:
\Pr(\Delta>0\mid\Delta\neq0)=\tfrac{1}{2},
\qquad
W\sim\operatorname{Binomial}(n,\tfrac{1}{2}).
\end{align}
The two-sided exact \(p\)-value is
\begin{align}
p_{\mathrm{sign}}
=
\min\left\{
1,\;
2\sum_{r=0}^{\min(W,L)}
\binom{n}{r}2^{-n}
\right\}.
\label{eq:sign-test}
\end{align}

Counting directly from Table~1 gives the following results.

\begin{table}[h]
\centering
\setlength{\tabcolsep}{3pt}
\begin{tabular}{lrrrr}
\hline
Full integration & \(W\) & \(L\) & \(T\) & \(p_{\mathrm{sign}}\)\\
\hline
GVM + Ours & 7 & 0 & 1 & 0.0156\\
PCL + Ours & 7 & 1 & 0 & 0.0703\\
GRESO + Ours & 7 & 0 & 1 & 0.0156\\
\hline
Pooled & 21 & 1 & 2 & \(1.10\times10^{-5}\)\\
\hline
\end{tabular}
\caption{Directional sign-test results for the full integrations. Ties are
excluded from the exact test.}
\label{tab:directional-sign-test}
\end{table}

For the PCL integration, \(W=7\), \(L=1\), and \(n=8\). Its two-sided exact
\(p\)-value is
\begin{align}
p_{\mathrm{sign}}
&=
2\left[
\binom{8}{0}+\binom{8}{1}
\right]2^{-8}\nonumber\\
&=
\frac{18}{256}
=
0.0703125.
\end{align}

At the nominal significance level \(\alpha=0.10\), all three full
integrations exhibit statistically significant directional improvements.
Specifically, GVM + Ours and GRESO + Ours achieve significance at the
\(\alpha=0.05\) level, while PCL + Ours achieves significance at the
\(\alpha=0.10\) level. These results demonstrate that the positive effect of
our method is not confined to a particular scheduler, base model, or
benchmark, but is consistently observed across the evaluated configurations.

For the pooled analysis, \(W=21\), \(L=1\), and \(n=22\). Equation~
\eqref{eq:sign-test} therefore gives
\begin{align}
p_{\mathrm{sign}}
&=
2\left[
\binom{22}{0}+\binom{22}{1}
\right]2^{-22}\nonumber\\
&=
\frac{46}{4{,}194{,}304}
=
1.0967\times10^{-5}.
\end{align}

The pooled result provides strong evidence against the null hypothesis that
positive and negative changes are equally likely. Across the three full
integrations, our method improves 21 of the 22 non-tied experimental
comparisons, corresponding to a gain rate of \(95.5\%\). The exact sign test
therefore confirms a highly consistent positive improvement direction across
the evaluated schedulers, base models, and benchmarks. This cross-setting
consistency is particularly valuable because it shows that the benefit of our
method generalizes across different integration strategies and experimental
conditions rather than depending on a single favorable configuration.

\section{D. Cross-Domain Generalization to Code}
\begin{table}[h]
\centering
\begin{tabular}{cccccc}
\hline
\multirow{2}{*}{Base Model} & \multirow{2}{*}{Methods} & \multicolumn{2}{c}{Full Steps} \\ \cline{3-4} 
 &  & MAE (↓) & r (↑) \\ \hline
\multirow{4}{*}{\begin{tabular}[c]{@{}c@{}}Qwen2.5-Coder-7B\\ Instruct\end{tabular}} & VIP & 0.197 & 0.504 \\
 & PCL & 0.402 & 0.307 \\
 & MoPPS & 0.224 & 0.773 \\
 & Ours & {\underline{0.176}} & {\underline{0.845}} \\ \hline
\end{tabular}
\caption{Difficulty-estimation performance on code generation using Qwen2.5-Coder-7B-Instruct. MAE is sample-level, and $r$ is the batch-level Pearson correlation.}
\label{tab:code_gen}
\end{table}
\subsection{Setting}
Many existing adaptive RL scheduling methods are developed and evaluated primarily for mathematical reasoning. Extending them to code generation may require domain-specific modifications to reward definitions, sampling strategies, or hyperparameters. Reimplementing these adaptations may result in unequal levels of optimization across baselines and compromise the fairness of the comparison. Therefore, rather than reproducing all scheduling baselines in the code domain, we examine whether our difficulty estimator can be directly transferred to code generation and track changes in sample difficulty during RL training.

We conduct RL training on LiveCodeBench~\cite{jain2024livecodebench} release\_v2, which contains 511 programming problems, using Qwen2.5-Coder-7B-Instruct. LiveCodeBench collects problems from LeetCode, AtCoder, and Codeforces and provides platform-derived difficulty ratings and executable test cases \citep{jain2024livecodebench}. A generated program is considered correct only if it passes all corresponding test cases.

Consistent with the main experiments, we generate eight rollouts for each selected sample and use the proportion of correct programs as its empirical success rate. We then replay the estimation process over the complete training trajectory. Before each training step, the estimator predicts sample success probabilities using only rollout feedback available from previous steps, and these predictions are compared with the subsequently observed outcomes.

\subsection{Results}
As shown in Table~\ref{tab:code_gen}, our method consistently tracks the evolving success probabilities of training samples throughout the RL trajectory of Qwen2.5-Coder-7B-Instruct. This result indicates that the proposed estimator does not depend on the exact-match reward commonly used in mathematical reasoning tasks. When the feedback is replaced by program correctness determined through test-case execution, the sample graph and historical rollout outcomes still provide effective information for estimating dynamic sample difficulty.

Overall, the experiment provides preliminary evidence that our estimator can be transferred to execution-based code generation without modifications specific to the code domain.

\subsection{Discussion}

The practical need for our estimator may be weaker in code generation. Some competitive-programming benchmarks provide platform difficulty ratings and execution feedback, which may already suffice for curriculum learning or rollout allocation in certain settings. This may partly explain the stronger focus of related studies on mathematics. Thus, this experiment establishes transferability rather than a clear advantage in code generation.

\section{E. Hyper Parameter Analysis}
\subsection{Guideline of $K$, $k_{knn}$ and $\beta$}
We discuss three hyperparameters controlling our difficulty estimator. $K$ is the number of latent difficulty states, determining modeling granularity; $k_{\mathrm{nn}}$ is the neighborhood size for graph construction, controlling feedback propagation range; $\beta$ is the Potts prior strength, balancing graph smoothing against adaptation to policy changes.

Firstly, we analyze the influence of hyperparameters $\beta$ and $K$ on difficulty estimation, as shown in Table~\ref{tab:beta_k_ablation}, where MAE is sample-level and $r$ is the batch-level Pearson correlation. When $\beta$ increases from $2$ to $5$ or above, all metrics become nearly identical and slightly worse, indicating that an over-strong Potts prior makes the model rely too heavily on neighbor information and neglect the global difficulty structure. In contrast, a smaller $K$ slightly improves the overall correlation $r$, suggesting that fewer latent states better capture the global difficulty trend, but the sample-level MAE increases, degrading per-sample accuracy. Considering this trade-off, we choose $\beta=2$ and $K=320$ in the main experiments.

\begin{table}[h]
\centering
\begin{tabular}{cccccc}
\hline
\multirow{2}{*}{$\beta$} & \multirow{2}{*}{K} & \multicolumn{2}{l}{Early Steps} & \multicolumn{2}{l}{Full Steps} \\
 &  & r (↑) & MAE (↓) & r (↑) & MAE (↓) \\ \hline
2 & 320 & 0.482 & 0.290 & 0.836 & 0.183 \\
2 & 160 & 0.480 & 0.306 & 0.838 & 0.191 \\
2 & 80 & 0.494 & 0.313 & 0.843 & 0.194 \\
2 & 40 & 0.507 & 0.327 & 0.848 & 0.196 \\
1 & 320 & 0.488 & 0.302 & 0.849 & 0.189 \\
5 & 320 & 0.494 & 0.297 & 0.847 & 0.186 \\
10 & 320 & 0.490 & 0.297 & 0.847 & 0.186 \\
20 & 320 & 0.493 & 0.298 & 0.847 & 0.187 \\ \hline
\end{tabular}
\caption{Sensitivity analysis of $\beta$ and $K$ on Qwen-2.5-Math-1.5B. MAE is sample-level, and $r$ is batch-level Pearson correlation.}
\label{tab:beta_k_ablation}
\end{table}

The neighborhood size \(k_{\mathrm{nn}}\) determines the initial connectivity of the difficulty-aware sample graph, and thereby controls how far rollout feedback can propagate across samples. Experiments show that as long as \(k_{\mathrm{nn}}\) is not set extremely small (e.g., below 10), the estimation accuracy is barely affected: after reciprocal-neighbor and positive-similarity filtering, the graph itself is already sufficiently sparse, so a small number of neighbors suffices for effective feedback propagation. Meanwhile, the ablation study shows that removing graph sparsification leads to only a modest performance drop, indicating that excessively increasing the neighborhood size brings no additional benefit. From a computational perspective, increasing \(k_{\mathrm{nn}}\) raises the cost of message propagation on the graph---in each variational inference iteration, updating the latent-state distribution of each sample requires aggregating information from its neighbors, giving a theoretical complexity of \(O(N k_{\mathrm{nn}})\). However, due to random memory access on sparse graphs and edge saturation caused by positive-similarity filtering, the actual wall-clock time does not grow linearly with \(k_{\mathrm{nn}}\)\footnote{Theoretically, the complexity grows linearly with \(k_{\mathrm{nn}}\), but in practice, graph sparsity leads to random memory access, and positive-similarity filtering gradually saturates the actual edge count, so the time does not increase linearly even if \(k_{\mathrm{nn}}\) is increased. In our setting, the wall-clock time when the edge count saturates is about \(1.61\times\) that of the proposed method in the main text.}.
\subsection{Discussion of Embedding Models and Difficulty-Aware Instruction}

\subsubsection{Models Selection}

We compare several recent text-embedding models that support task-aware or instruction-conditioned embedding, including Qwen3-Embedding-0.6B, Qwen3-Embedding-4B~\cite{zhang2025qwen3embeddingadvancingtext}, and BGE-M3~\cite{chen2024bgem3}. When their standard dense representations are used to construct the sample graph, the resulting performance is broadly similar (see Table~\ref{tab:embedding_ablation}), suggesting that our framework is not sensitive to the exact choice among recent embedding models. Among the direct, uncompressed configurations, Qwen3-Embedding-0.6B performs on par with the larger Qwen3-Embedding-4B and BGE-M3 on Qwen-2.5-Math-1.5B (0.183 vs.\ 0.188 vs.\ 0.188 MAE), while on Llama-3.2-1B-Instruct it is slightly worse than Qwen3-Embedding-4B (0.131 vs.\ 0.130). Given that Qwen3-Embedding-0.6B is substantially more efficient, we adopt it as the default encoder as a trade-off between accuracy and computational cost.

Model scale and raw embedding dimensionality do not exhibit a monotonic relationship with downstream performance. In particular, compressing the Qwen3-Embedding-4B output to 1,024 dimensions using its Matryoshka-compatible representation~\cite{kusupati2022matryoshka} improves its performance relative to using the full-dimensional representation (0.179 vs.\ 0.188 MAE on Qwen; 0.121 vs.\ 0.130 on Llama, as shown in Table~\ref{tab:embedding_ablation}). Notably, the MRL-compressed Qwen3-Embedding-4B at 1,024 dimensions achieves the best overall results on both base models (0.179/0.844 for Qwen and 0.121/0.787 for Llama), suggesting that a larger encoder combined with dimensionality reduction can further improve graph quality if the additional computational cost is acceptable. Nevertheless, for the main experiments we prioritize low overhead and therefore use Qwen3-Embedding-0.6B as the default.

A plausible explanation for the benefit of MRL compression is the curse of dimensionality: in an excessively high-dimensional space, pairwise distances can become less discriminative, which may make nearest-neighbor selection noisier and weaken the local graph structure. Matryoshka Representation Learning (MRL) preserves information at nested dimensionalities, allowing the representation to be shortened without applying an arbitrary post-hoc projection. Our observation is consistent with this explanation, although it does not by itself establish dimensionality as the sole causal factor.

In contrast, replacing the recent instruction-aware encoders with earlier Sentence-BERT-family models~\cite{reimers2019sentencebert}, or with the encoder configuration adopted by VIP~\cite{nguyen2026adaptivevip}, causes a substantial performance drop. These older representations were primarily optimized for general sentence-level similarity and appear less suitable for capturing the task-specific semantic, reasoning, and difficulty cues required by our graph. Based on these results, we recommend pairing our method with a recent task-aware embedding model. Qwen3-Embedding-0.6B provides a strong default efficiency--performance trade-off, while a larger MRL-compatible encoder truncated to a moderate dimension is a viable alternative.

\subsubsection{Discussion of Difficulty-Aware Instruction}

\begin{table}[h]
\small
\begin{tabular}{cccc}
\hline
\multirow{2}{*}{Base Model} & \multirow{2}{*}{Methods} & \multicolumn{2}{c}{Full Steps} \\
 &  & MAE (↓) & r (↑) \\ \hline
\multirow{6}{*}{\begin{tabular}[c]{@{}c@{}}Qwen-2.5\\ Math-1.5B\end{tabular}} & Qwen3-Embedding-0.6B & 0.183 & 0.836 \\
 & -Difficulty-Aware Instruction & 0.309 & 0.583 \\
 & Qwen3-Embedding-4B & 0.188 & 0.827 \\
 & \begin{tabular}[c]{@{}c@{}}Qwen3-Embedding-4B\\ MRL(1024 dimensions)\end{tabular} & 0.179 & 0.844 \\
 & BGE-M3 & 0.188 & 0.823 \\
 & MiniLM & 0.326 & 0.407 \\ \hline
\multirow{6}{*}{\begin{tabular}[c]{@{}c@{}}Llama-3.2\\ 1B-Instruct\end{tabular}} & Qwen3-Embedding-0.6B & 0.131 & 0.776 \\
 & -Difficulty-Aware Instruction & 0.302 & 0.486 \\
 & Qwen3-Embedding-4B & 0.130 & 0.749 \\
 & \begin{tabular}[c]{@{}c@{}}Qwen3-Embedding-4B\\ MRL(1024 dimensions)\end{tabular} & 0.121 & 0.787 \\
 & BGE-M3 & 0.139 & 0.783 \\
 & MiniLM & 0.304 & 0.407 \\ \hline
\end{tabular}
\caption{Full-trajectory difficulty-estimation performance of different embedding configurations. The row ``-Difficulty-Aware Instruction'' removes the instruction from the default encoder. MAE: sample-level mean absolute error (lower is better); $r$: batch-level Pearson correlation (higher is better).}
\label{tab:embedding_ablation}
\end{table}

Table~\ref{tab:embedding_ablation} shows that removing the difficulty-aware instruction from Qwen3-Embedding-0.6B causes a substantial degradation in estimation quality. On Qwen-2.5-Math-1.5B, the MAE increases from 0.183 to 0.309 and the correlation drops from 0.836 to 0.583; on Llama-3.2-1B-Instruct, the MAE increases from 0.131 to 0.302 and the correlation drops from 0.776 to 0.486. This confirms that the instruction, which encourages the encoder to capture difficulty-related and reasoning-oriented features beyond pure surface semantics, contributes critically to the graph quality. Without it, the embeddings tend to connect samples that are semantically similar but not necessarily informative for difficulty transfer.

The comparison between Qwen3-Embedding-4B and its MRL-truncated 1,024-dimension variant also suggests that reducing dimensionality, when done through a representation-learning-compatible method, can improve the discriminability of nearest-neighbor structure. BGE-M3 achieves comparable performance to the Qwen3 variants, while MiniLM performs poorly, which is consistent with the above observation that older generic encoders lack the task-awareness needed for constructing a difficulty-aware graph.

\subsection{Discussion of the Interpolation Weight $\gamma$}

For samples with prior rollouts, we combine the graph-based prediction and the sample-level empirical success rate as$\widehat p_{i,t}=\gamma\widehat p_{i,t}^{\mathrm{model}}+(1-\gamma)\bar r_{i,<t}$
The parameter $\gamma$ controls the balance between transferable graph-based evidence and direct sample-level history. A small $\gamma$ makes the estimate dominated by $(\bar r_{i,<t}$, which weakens graph propagation and can hurt difficulty ranking. A large $\gamma$ relies more on the latent-state model and may ignore reliable direct observations, leading to larger estimation error.

We set $\gamma=0.5$ as a general default. In our experiments, performance is relatively stable for $\gamma\in[0.5,0.8]$. When $\gamma<0.5$, ranking quality tends to decrease; when $\gamma>0.8$, sample-level estimation error usually increases. This suggests that a moderate interpolation weight is preferable.

A more principled choice would adapt $\gamma$ according to uncertainty. The uncertainty of latent state (k) can be estimated from the variance of its Beta posterior:
\[
v_{k,t-1}=\frac{a_{k,t-1}b_{k,t-1}}{(a_{k,t-1}+b_{k,t-1})^2(a_{k,t-1}+b_{k,t-1}+1)}.
\]
Combined with assignment uncertainty, this gives
\[
\widehat u_{i,t}=\sum_{k=1}^{K}q_{ik,t-1}\left[v_{k,t-1}+\left(\mu_{k,t-1}-\widehat p_{i,t}^{\mathrm{model}}\right)^2\right].
\]
In principle, one could increase $\gamma$ when the model-based prediction is confident, and decrease it when the sample has sufficient direct observations. However, such a rule also depends on rollout counts, sampling distributions, and the downstream scheduler. These factors vary across integrated frameworks and are hard to fix with limited experiments. We therefore use a fixed \(\gamma=0.5\).

\section{F. Discussion of Spectral Clusters}

\subsection{Association with Subject Categories}

For each analyzed MATH problem \(i\), let \(z_i\in\{1,\ldots,K_+\}\) denote
its nonempty spectral-cluster assignment and let
\(c_i\in\{1,\ldots,U\}\) denote its subject category. We form the
\(K_+\times U\) contingency table
\begin{align}
O_{ku}=\sum_{i=1}^{n}\mathbb{I}(z_i=k,c_i=u),
\end{align}
with row totals \(O_{k\cdot}\), column totals \(O_{\cdot u}\), and total
sample size \(n\). Under the null hypothesis that cluster assignment and
subject category are independent, the expected count in cell \((k,u)\) is
\begin{align}
E_{ku}=\frac{O_{k\cdot}O_{\cdot u}}{n}.
\end{align}
We measure departure from independence using Pearson's chi-squared
statistic~\cite{pearson1900criterion},
\begin{align}
\chi^2
=\sum_{k=1}^{K_+}\sum_{u=1}^{U}
\frac{(O_{ku}-E_{ku})^2}{E_{ku}},
\end{align}
whose asymptotic null distribution has
\((K_+-1)(U-1)\) degrees of freedom. Applying this calculation to the
MATH training split gives \(n=7{,}500\), \(K_+=16\), and \(U=7\);
thus, the contingency table contains \(16\times7\) cells and the asymptotic
reference distribution has \(90\) degrees of freedom. The resulting statistic
is \(\chi^2=10{,}995.84\), with asymptotic \(p<10^{-300}\), rejecting
independence.

Because the chi-squared statistic increases with sample size, we additionally
report Cramér's \(V\), a normalized effect-size measure for categorical
association~\cite{cramer1946mathematical}:
\begin{align}
V
=\sqrt{
\frac{\chi^2}
{n\min(K_+-1,U-1)}
}.
\end{align}
The resulting \(V=0.494\) indicates a substantial association between the
spectral clusters and the annotated mathematical subjects. This effect-size
analysis complements the hypothesis test: the \(p\)-value establishes that
the association is unlikely under independence, whereas \(V\) quantifies its
magnitude.

\subsection{Difficulty Differences Within Subject Categories}

Subject category and cluster assignment are strongly associated, so a single
pooled comparison could confound subject composition with difficulty. We
therefore conduct seven separate Kruskal--Wallis tests, one within each MATH
subject category~\cite{kruskal1952ranks}. This stratification asks whether
the annotated difficulty levels differ across spectral clusters among
problems from the same subject.

For subject \(u\), let \(n_u\) be its number of problems and \(n_{ku}\) the
number assigned to nonempty cluster \(k\). We rank the ordinal difficulty
levels of all \(n_u\) problems within that subject, assigning midranks to
ties. Let \(R_{ku}\) be the sum of ranks in cluster \(k\). The uncorrected
Kruskal--Wallis statistic is
\begin{align}
H_u^{(0)}
=\frac{12}{n_u(n_u+1)}
\sum_{k:n_{ku}>0}\frac{R_{ku}^{2}}{n_{ku}}
-3(n_u+1).
\end{align}
Because MATH difficulty takes only five ordinal levels, ties are frequent.
Let \(t_{ug}\) be the size of the \(g\)-th tied group within subject \(u\).
We apply the standard ties correction
\begin{align}
C_u
=1-\frac{\sum_g(t_{ug}^{3}-t_{ug})}{n_u^{3}-n_u},
\qquad
H_u=\frac{H_u^{(0)}}{C_u}.
\end{align}
If \(K_u\) clusters are nonempty within subject \(u\), the raw asymptotic
\(p\)-value is computed from a chi-squared reference distribution with
\(K_u-1\) degrees of freedom:
\begin{align}
p_u=\Pr\!\left(\chi^2_{K_u-1}\geq H_u\right).
\end{align}

The seven raw \(p\)-values form one family of hypotheses. We therefore apply
the Benjamini--Hochberg procedure to control the false discovery
rate~\cite{benjamini1995fdr}. After sorting the raw values as
\(p_{(1)}\leq\cdots\leq p_{(m)}\), where \(m=7\), the monotone adjusted values
are
\begin{align}
\widetilde p_{(r)}
=\min\left\{
1,\;
\min_{j\geq r}\frac{m}{j}p_{(j)}
\right\}.
\end{align}

\begin{table}[h]
\centering
\begin{tabular}{lcc}
\hline
Subject & Raw \(p\) & BH-adjusted \(p\)\\
\hline
Algebra & \(\approx0\) & \(\approx0\)\\
Prealgebra & \(\approx0\) & \(\approx0\)\\
Precalculus & \(\approx0\) & \(1.0\times10^{-6}\)\\
Intermediate algebra & \(2.0\times10^{-6}\) & \(3.0\times10^{-6}\)\\
Counting/probability & \(8.0\times10^{-6}\) & \(1.1\times10^{-5}\)\\
Number theory & \(3.19\times10^{-4}\) & \(3.72\times10^{-4}\)\\
Geometry & \(4.31\times10^{-3}\) & \(4.31\times10^{-3}\)\\
\hline
\end{tabular}
\caption{Within-subject Kruskal--Wallis tests of difficulty differences across
spectral clusters. Values displayed as zero are below the reporting precision;
BH correction uses the unrounded raw values.}
\label{tab:within-subject-kw}
\end{table}

All seven subject-specific null hypotheses remain rejected after BH
correction, with \(\widetilde p_u\leq4.31\times10^{-3}\). Therefore, cluster
membership is associated with annotated difficulty even among problems from
the same subject category. Together with the cluster--subject association,
this result supports the interpretation that the spectral initialization
captures both semantic and difficulty-related structure.

\section{G. Windowed Memory under Non-stationary Policies}

\subsection{Windowed Forgetting Mechanism}

The online inference procedure described in the main text accumulates rollout feedback across training steps by inheriting the state-level Beta posterior, i.e., \(a_{k,t}^{\mathrm{hist}}=a_{k,t-1}\) and \(b_{k,t}^{\mathrm{hist}}=b_{k,t-1}\). However, the rollout distribution is non-stationary because the policy parameters continuously change during RL optimization. Consequently, rollout observations collected many training steps earlier may no longer accurately characterize the success probability of the current policy. To reduce the influence of such outdated observations while remaining responsive to the current policy, we introduce a windowed forgetting mechanism. The windowed variant preserves the difficulty-aware sample graph \(W\), the smoothed assignment prior \(A\), the Potts graph prior, and the mean-field variational family introduced in the main method. It only changes how historical rollout evidence is organized: state-level success probabilities are estimated from a recent sliding window, whereas the complete rollout history of each previously observed sample is retained for sample-level prediction.

\subsubsection{Windowed State-Level Statistics.}

Before predicting training step \(t\), we define the preceding window of length \(L\geq 1\) as
\begin{align}
\mathcal{T}_{t}^{(L)}=\left\{\tau:\max(1,t-L)\leq \tau<t\right\},
\end{align}
where \(L\) denotes the window length and is distinguished from the graph adjacency matrix \(W\). For latent state \(k\), the responsibility-weighted numbers of successful and unsuccessful rollouts within this window are
\begin{align}
S_{k,t}^{(L)}=\sum_{\tau\in\mathcal{T}_{t}^{(L)}}\sum_{i\in\mathcal{O}_{\tau}}q_{ik,\tau}s_{i,\tau},
\end{align}
and
\begin{align}
F_{k,t}^{(L)}=\sum_{\tau\in\mathcal{T}_{t}^{(L)}}\sum_{i\in\mathcal{O}_{\tau}}q_{ik,\tau}\left(n_{i,\tau}-s_{i,\tau}\right).
\end{align}
Rather than directly inheriting the complete Beta posterior from the previous step, the windowed variant reconstructs the state-level Beta parameters from a fixed prior and the rollout evidence within the current window:
\begin{align}
a_{k,t}^{(L)}=a_0+S_{k,t}^{(L)},\qquad b_{k,t}^{(L)}=b_0+F_{k,t}^{(L)}.
\end{align}
In our implementation, we set \(a_0=b_0=1\), corresponding to the uniform prior \(\operatorname{Beta}(1,1)\). This is a standard weak prior for a Bernoulli success probability and does not favor any particular success rate before rollout evidence is observed. It also guarantees that the Beta parameters remain strictly positive. This property is particularly important under a finite window because some latent states may receive no rollout observations in the current window. When \(S_{k,t}^{(L)}=F_{k,t}^{(L)}=0\), the \(\operatorname{Beta}(1,1)\) prior still yields a valid posterior and keeps the posterior mean and the digamma terms used in the variational updates well-defined and numerically stable.

Ideally, the prior mean
\begin{align}
\mathbb{E}[\theta_{k,t}]=\frac{a_0}{a_0+b_0}
\end{align}
should be close to the average success probability of the current policy. Such a prior would provide a more informative default estimate for latent states with few or no recent observations. However, obtaining a reliable estimate of the current policy's average ability would require evaluating a sufficiently large and representative subset of the training set at each stage, which would introduce substantial additional rollout cost. Estimating it only from the samples selected by the scheduler would also be potentially biased, because the selected samples are generally not representative of the full training distribution. We therefore adopt \(\operatorname{Beta}(1,1)\) as a practical trade-off among prior neutrality, numerical stability, and computational efficiency.

The posterior mean success probability of latent state \(k\) is then
\begin{align}
\mu_{k,t}^{(L)}=\frac{a_{k,t}^{(L)}}{a_{k,t}^{(L)}+b_{k,t}^{(L)}}.
\end{align}
The model-based prediction for sample \(x_i\) retains the same mixture form as in the main method:
\begin{align}
\widehat{p}_{i,t}^{\mathrm{model}}=\sum_{k=1}^{K}q_{ik,t}\mu_{k,t}^{(L)}.
\end{align}
When the window covers all preceding training steps, the windowed formulation reduces to the cumulative-history variant. With a finite \(L\), rollout observations older than \(L\) training steps no longer directly contribute to the current state-level success probabilities.
\subsubsection{Windowed Variational Updates.}
The windowed variant retains the mean-field factorization
\begin{align}
q_t(\mathbf{z}_t,\boldsymbol{\theta}_t)\approx\prod_{i=1}^{N}q_i(z_{i,t})\prod_{k=1}^{K}q(\theta_{k,t}),
\end{align}
where \(q_{ik,t}\equiv q_i(z_{i,t}=k)\) and
\begin{align}
q(\theta_{k,t})=\operatorname{Beta}\left(a_{k,t}^{(L)},b_{k,t}^{(L)}\right).
\end{align}
The coordinate update for the latent-state assignments preserves the same three sources of information as the main method:
$q_{ik}^{(r+1)}=\operatorname{Softmax}_{k}\left[\log A_{ik}+\beta\sum_{j=1}^{N}W_{ij}q_{jk}^{(r)}+\mathbb{I}\left(i\in\mathcal{O}_{t}^{(L)}\right)\ell_{ik,t}^{(L)}\right],$
where
\begin{align}
\mathcal{O}_{t}^{(L)}=\bigcup_{\tau\in\mathcal{T}_{t}^{(L)}}\mathcal{O}_{\tau}.
\end{align}
For each sample \(i\), we aggregate its rollout outcomes within the current window as
\begin{align}
s_{i,t}^{(L)}=\sum_{\tau\in\mathcal{T}_{t}^{(L)}}s_{i,\tau},\qquad n_{i,t}^{(L)}=\sum_{\tau\in\mathcal{T}_{t}^{(L)}}n_{i,\tau},
\end{align}
and compute the corresponding expected log-likelihood as
$\ell_{ik,t}^{(L)}={}s_{i,t}^{(L)}\left[\psi\left(a_{k,t}^{(L)}\right)-\psi\left(a_{k,t}^{(L)}+b_{k,t}^{(L)}\right)\right]+\left(n_{i,t}^{(L)}-s_{i,t}^{(L)}\right)\left[\psi\left(b_{k,t}^{(L)}\right)-\psi\left(a_{k,t}^{(L)}+b_{k,t}^{(L)}\right)\right].$

After updating the assignment probabilities, the state-level Beta factors are recomputed using only the rollout observations within the current window:
\begin{align}
a_{k,t}^{(L)}=1+\sum_{\tau\in\mathcal{T}_{t}^{(L)}}\sum_{i\in\mathcal{O}_{\tau}}q_{ik}s_{i,\tau},
\end{align}
and
\begin{align}
b_{k,t}^{(L)}=1+\sum_{\tau\in\mathcal{T}_{t}^{(L)}}\sum_{i\in\mathcal{O}_{\tau}}q_{ik}\left(n_{i,\tau}-s_{i,\tau}\right).
\end{align}
We alternate the assignment and Beta-factor updates using the same stopping criterion as in the main method. Therefore, the windowed variant does not change the coordinate-ascent structure of the inference procedure; it only restricts the temporal support of the rollout evidence used in the likelihood and state-level Beta updates.

\subsubsection{Persistent Assignment Warm Start.}

In the cumulative formulation, the complete estimator state \(\mathcal{S}_t=\{q_t,a_t,b_t\}\) is passed between consecutive training steps. In the windowed variant, we retain only the previous assignment probabilities as the default initialization:
\begin{align}
q_{ik,t}^{(0)}=q_{ik,t-1}.
\end{align}
The Beta parameters are not directly inherited from the previous step. Instead, they are reconstructed from the fixed \(\operatorname{Beta}(1,1)\) prior and the rollout observations within the current window. The persistent initialization of \(q\) preserves continuity in the graph-structured latent assignments and provides a warm start for the finite-iteration variational procedure. Importantly, the inherited responsibilities are used only as an optimization initialization. They are not treated as additional successes or failures and therefore do not alter the strict temporal window applied to the state-level sufficient statistics. We additionally consider a reset variant that initializes \(q\) from the static soft assignment prior \(A\) at every step; unless otherwise specified, the persistent warm-start variant is used by default.

\subsubsection{Persistent Sample-Level History.}

The sliding window is applied only to the shared state-level statistics. For each sample \(x_i\), we separately retain its complete rollout history before step \(t\):
\begin{align}
S_{i,t}^{\mathrm{all}}=\sum_{\tau<t}s_{i,\tau},\qquad C_{i,t}^{\mathrm{all}}=\sum_{\tau<t}n_{i,\tau}.
\end{align}
For a previously observed sample, its historical empirical success rate is
\begin{align}
\overline{r}_{i,<t}=\frac{S_{i,t}^{\mathrm{all}}}{C_{i,t}^{\mathrm{all}}}.
\end{align}
We then retain the prediction rule from the main method:
\begin{align}
\widehat{p}_{i,t}=\begin{cases}\widehat{p}_{i,t}^{\mathrm{model}},&C_{i,t}^{\mathrm{all}}=0,\\[4pt]\gamma\widehat{p}_{i,t}^{\mathrm{model}}+(1-\gamma)\overline{r}_{i,<t},&C_{i,t}^{\mathrm{all}}>0.\end{cases}
\end{align}
Consequently, the two components of the final prediction operate at complementary time scales. The model-based term uses recent rollout feedback to track the evolving success probabilities of the shared latent states, whereas the sample-level empirical term preserves all available direct evidence for each previously observed sample. The windowed mechanism therefore removes stale evidence from transferable state-level statistics without discarding the long-term historical experience associated with individual samples.

\subsection{Result and Discussion}
\begin{table}[h]
\centering
\begin{tabular}{cccc}
\hline
\multirow{2}{*}{Base Model} & \multirow{2}{*}{Methods} & \multicolumn{2}{c}{Full Steps} \\
 &  & MAE (↓) & r (↑) \\ \hline
\multirow{2}{*}{\begin{tabular}[c]{@{}c@{}}Qwen-2.5\\ Math-1.5B\end{tabular}} & Ours & 0.183 & 0.836 \\
 & Windows & 0.127 & 0.712 \\ \hline
\multirow{2}{*}{\begin{tabular}[c]{@{}c@{}}Llama-3.2\\ 1B-Instruct\end{tabular}} & Ours & 0.131 & 0.776 \\
 & Windows & 0.119 & 0.747 \\ \hline
\end{tabular}
\caption{Sample-level difficulty estimation performance of the cumulative estimator (Ours) and the windowed variant (Windows) over all training steps. MAE measures the absolute prediction error, while batch-level \(r\) denotes the Pearson correlation between the predicted and reference success probabilities. Lower MAE and higher \(r\) indicate better performance.}
\label{tab:forgetting_mae}
\end{table}
\begin{figure}[h]
    \centering
    \includegraphics[width=1\linewidth]{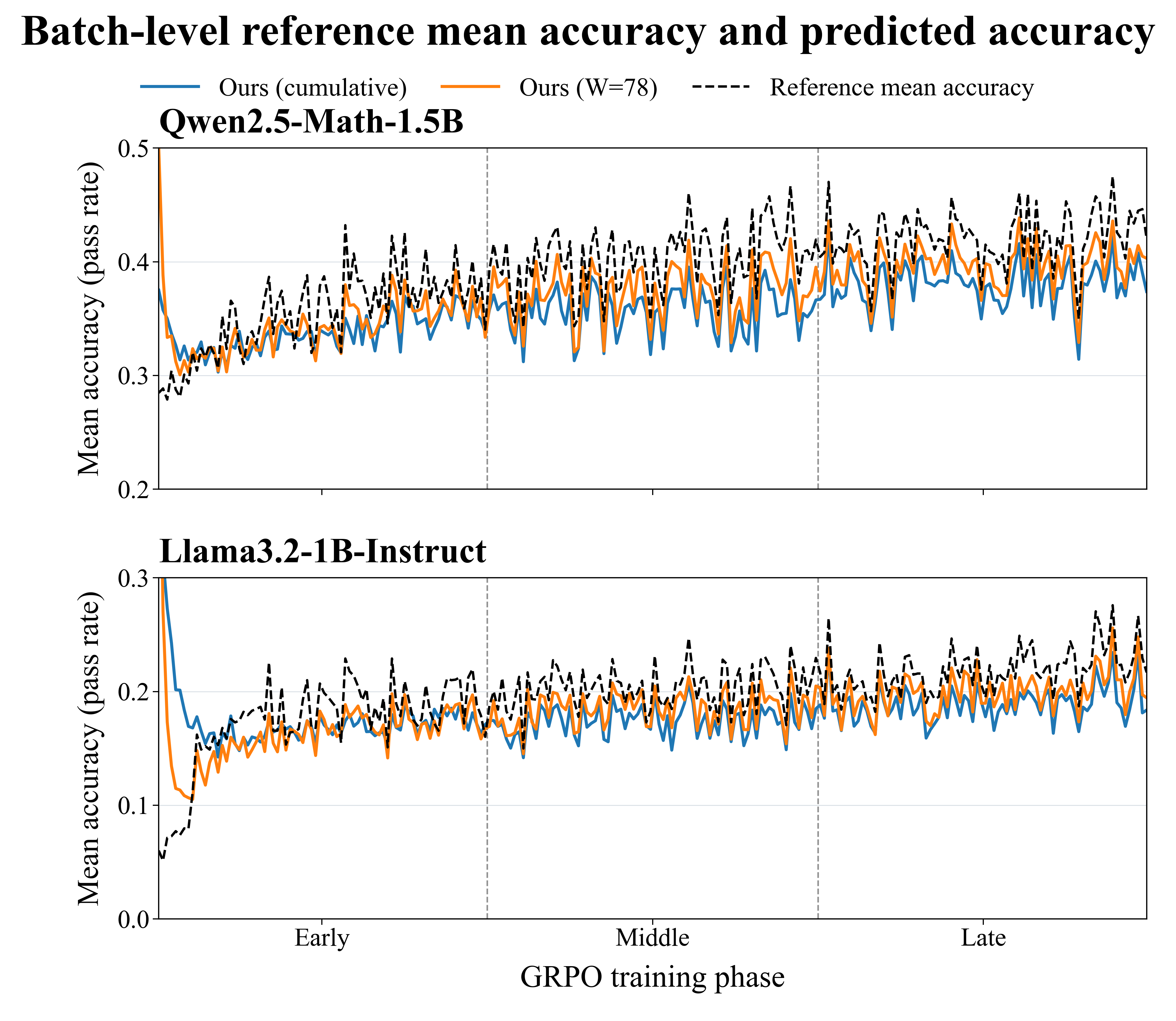}
    \caption{Batch-level reference accuracy and predicted accuracy of the cumulative and windowed estimators on Qwen-2.5-Math-1.5B and Llama-3.2-1B-Instruct.}
    \label{fig:forgetting_analysis}
\end{figure}

We evaluate the windowed variant by replaying the full training logs of Qwen-2.5-Math-1.5B and Llama-3.2-1B-Instruct, and compare it with the cumulative estimator used in our main experiments. As shown in Table~\ref{tab:forgetting_mae}, the windowed variant achieves lower sample-level MAE on both models: from \(0.183\) to \(0.127\) on Qwen-2.5-Math-1.5B, and from \(0.131\) to \(0.119\) on Llama-3.2-1B-Instruct. This indicates that windowed forgetting makes the predicted success probability of each individual sample numerically closer to its reference rollout outcome. Fig.~\ref{fig:forgetting_analysis} further visualizes the batch-level reference accuracy and predicted accuracy over the full training process.

However, Table~\ref{tab:forgetting_mae} also shows that the correlation metric \(r\) decreases after introducing the windowed mechanism: from \(0.836\) to \(0.712\) on Qwen-2.5-Math-1.5B, and from \(0.776\) to \(0.747\) on Llama-3.2-1B-Instruct. This suggests a trade-off: although the windowed estimator improves point-wise numerical accuracy, it weakens the ability to preserve the relative difficulty structure among samples. A likely reason is that the finite window reduces the amount of historical evidence available for state-level estimation, making the estimates more sensitive to sparse recent observations and sampling bias. For our difficulty-aware scheduling setting, stable relative ordering is more important than point-wise calibration alone. Therefore, we retain the cumulative estimator as the default method in the main experiments. Whether to use windowed forgetting should depend on the downstream objective: it is beneficial when current-policy calibration is prioritized, but less suitable when stable sample ranking is critical.

\section{H. Related work of Graph-Based Learning and Probabilistic Inference}
Graph-based clustering and semi-supervised learning assume that nearby points on a data manifold tend to have compatible latent structure. Spectral methods partition a similarity graph through its eigensystem~\cite{ng2001spectral,von2007tutorial}, while Gaussian-field propagation, local-and-global consistency, and manifold regularization diffuse sparse supervision according to graph geometry~\cite{zhu2003semi,NIPS2003_87682805,JMLR:v7:belkin06a}. These methods motivate transferring evidence from observed prompts to related but unobserved ones.
Probabilistic graphical models additionally represent relational dependence and uncertainty. Potts and Markov random-field priors encourage neighboring discrete variables to take compatible states~\cite{potts1952somepotts,besag1974spatialpotts,geman1984stochastic}. Because graph coupling makes exact posterior inference difficult, variational methods optimize a tractable approximation, with mean-field factorizations yielding scalable coordinate updates~\cite{zhang1992meanmrf,mcgrory2009variationalmrf,wainwright2008graphical,Blei03042017}. Stochastic and streaming variants update approximate posteriors as new observations arrive~\cite{hoffman2013stochastic,NIPS2013_51ef186e}.

\bibliography{aaai2027}

@article{guo2025deepseekr1,
  title={Deepseek-r1: Incentivizing reasoning capability in llms via reinforcement learning},
  author={Guo, Daya and Yang, Dejian and Zhang, Haowei and Song, Junxiao and Wang, Peiyi and Zhu, Qihao and Xu, Runxin and Zhang, Ruoyu and Ma, Shirong and Bi, Xiao and others},
  journal={arXiv preprint arXiv:2501.12948},
  year={2025}
}

@article{yu2026dapo,
  title={Dapo: An open-source llm reinforcement learning system at scale},
  author={Yu, Qiying and Zhang, Zheng and Zhu, Ruofei and Yuan, Yufeng and Zuo, Xiaochen and Yue, Yu and Dai, Weinan and Fan, Tiantian and Liu, Gaohong and Liu, Lingjun and others},
  journal={Advances in Neural Information Processing Systems},
  volume={38},
  pages={113222--113244},
  year={2026}
}

@inproceedings{verifystepbystep,
  title={Let's verify step by step},
  author={Lightman, Hunter and Kosaraju, Vineet and Burda, Yuri and Edwards, Harrison and Baker, Bowen and Lee, Teddy and Leike, Jan and Schulman, John and Sutskever, Ilya and Cobbe, Karl},
  booktitle={International Conference on Learning Representations},
  volume={2024},
  pages={39578--39601},
  year={2024}
}

@article{zheng2025groupgspo,
  title={Group sequence policy optimization},
  author={Zheng, Chujie and Liu, Shixuan and Li, Mingze and Chen, Xiong-Hui and Yu, Bowen and Gao, Chang and Dang, Kai and Liu, Yuqiong and Men, Rui and Yang, An and others},
  journal={arXiv preprint arXiv:2507.18071},
  year={2025}
}

@article{shao2024deepseekmath,
  title={Deepseekmath: Pushing the limits of mathematical reasoning in open language models},
  author={Shao, Zhihong and Wang, Peiyi and Zhu, Qihao and Xu, Runxin and Song, Junxiao and Bi, Xiao and Zhang, Haowei and Zhang, Mingchuan and Li, YK and Wu, Yang and others},
  journal={arXiv preprint arXiv:2402.03300},
  year={2024}
}

@inproceedings{
pcl,
title={Prompt Curriculum Learning for Efficient {LLM} Post-Training},
author={Zhaolin Gao and Joongwon Kim and Wen Sun and Thorsten Joachims and Sid Wang and Richard Yuanzhe Pang and Liang Tan},
booktitle={The Fourteenth International Conference on Learning Representations},
year={2026},
url={https://openreview.net/forum?id=zqOCacBD3P}
}

@article{zheng2026actgreso,
  title={Act only when it pays: Efficient reinforcement learning for llm reasoning via selective rollouts},
  author={Zheng, Haizhong and Zhou, Yang and Bartoldson, Brian and Kailkhura, Bhavya and Lai, Fan and Zhao, Jiawei and Chen, Beidi},
  journal={Advances in Neural Information Processing Systems},
  volume={38},
  pages={124321--124346},
  year={2026}
}

@article{yao2026optimizinggvm,
  title={Optimizing chain-of-thought reasoners via gradient variance minimization in rejection sampling and rl},
  author={Yao, Jiarui and Hao, Yifan and Zhang, Hanning and Dong, Hanze and Xiong, Wei and Jiang, Nan and Zhang, Tong},
  journal={Advances in Neural Information Processing Systems},
  volume={38},
  pages={163245--163284},
  year={2026}
}

@inproceedings{
zeng2026cures,
title={Cur{ES}: From Gradient Analysis to Efficient Curriculum Learning for Reasoning {LLM}s},
author={Yongcheng Zeng and Zexu Sun and Bokai Ji and Erxue Min and Hengyi Cai and Shuaiqiang Wang and Dawei Yin and Haifeng Zhang and Xu Chen and Jun Wang},
booktitle={The Fourteenth International Conference on Learning Representations},
year={2026},
url={https://openreview.net/forum?id=QXrZ0Y3yGJ}
}

@inproceedings{
nguyen2026adaptivevip,
title={Adaptive Rollout Allocation for Online Reinforcement Learning with Verifiable Rewards},
author={Hieu Trung Nguyen and Bao Nguyen and Wenao Ma and Yuzhi Zhao and Ruifeng She and Viet Anh Nguyen},
booktitle={The Fourteenth International Conference on Learning Representations},
year={2026},
url={https://openreview.net/forum?id=Z5sWYACAop}
}

@inproceedings{cdas,
  title={Rethinking the sampling criteria in reinforcement learning for LLM reasoning: A competence-difficulty alignment perspective},
  author={Kong, Deyang and Guo, Qi and Xi, Xiangyu and Wang, Wei and Wang, Jingang and Cai, Xunliang and Zhang, Shikun and Ye, Wei},
  booktitle={Proceedings of the AAAI Conference on Artificial Intelligence},
  volume={40},
  number={37},
  pages={31438--31446},
  year={2026}
}

@inproceedings{nguyen2026space,
  title={SPaCe: Unlocking Sample-Efficient Large Language Models Training With Self-Pace Curriculum Learning},
  author={Nguyen, Manh and Venkatesh, Svetha and Le, Hung and others},
  booktitle={Findings of the Association for Computational Linguistics: ACL 2026},
  pages={3480--3507},
  year={2026}
}

@inproceedings{qu2026can,
  title={Can prompt difficulty be online predicted for accelerating rl finetuning of reasoning models?},
  author={Qu, Yun and Wang, Qi and Mao, Yixiu and Hu, Vincent Tao and Ommer, Bj{\"o}rn and Ji, Xiangyang},
  booktitle={Proceedings of the 32nd ACM SIGKDD Conference on Knowledge Discovery and Data Mining V. 1},
  pages={1240--1250},
  year={2026}
}

@article{team2025kimi,
  title={Kimi k1. 5: Scaling reinforcement learning with llms},
  author={Team, Kimi and Du, Angang and Gao, Bofei and Xing, Bowei and Jiang, Changjiu and Chen, Cheng and Li, Cheng and Xiao, Chenjun and Du, Chenzhuang and Liao, Chonghua and others},
  journal={arXiv preprint arXiv:2501.12599},
  year={2025}
}

@article{li2024numinamath,
  title={Numinamath: The largest public dataset in ai4maths with 860k pairs of competition math problems and solutions},
  author={Li, Jia and Beeching, Edward and Tunstall, Lewis and Lipkin, Ben and Soletskyi, Roman and Huang, Shengyi and Rasul, Kashif and Yu, Longhui and Jiang, Albert Q and Shen, Ziju and others},
  journal={Hugging Face repository},
  volume={13},
  number={9},
  pages={9},
  year={2024}
}

@article{cui2025process,
  title={Process reinforcement through implicit rewards},
  author={Cui, Ganqu and Yuan, Lifan and Wang, Zefan and Wang, Hanbin and Zhang, Yuchen and Chen, Jiacheng and Li, Wendi and He, Bingxiang and Fan, Yuchen and Yu, Tianyu and others},
  journal={arXiv preprint arXiv:2502.01456},
  year={2025}
}

@article{hendrycks2021measuringmath,
  title={Measuring mathematical problem solving with the math dataset},
  author={Hendrycks, Dan and Burns, Collin and Kadavath, Saurav and Arora, Akul and Basart, Steven and Tang, Eric and Song, Dawn and Steinhardt, Jacob},
  journal={arXiv preprint arXiv:2103.03874},
  year={2021}
}

@article{balunovic2026matharenaaime2024,
  title={Matharena: Evaluating llms on uncontaminated math competitions},
  author={Balunovic, Mislav and Dekoninck, Jasper and Petrov, Ivo and Jovanovi{\'c}, Nikola and Vechev, Martin},
  journal={Advances in Neural Information Processing Systems},
  volume={38},
  year={2026}
}

@inproceedings{he2024olympiadbench,
  title={Olympiadbench: A challenging benchmark for promoting agi with olympiad-level bilingual multimodal scientific problems},
  author={He, Chaoqun and Luo, Renjie and Bai, Yuzhuo and Hu, Shengding and Thai, Zhen and Shen, Junhao and Hu, Jinyi and Han, Xu and Huang, Yujie and Zhang, Yuxiang and others},
  booktitle={Proceedings of the 62nd Annual Meeting of the Association for Computational Linguistics (Volume 1: Long Papers)},
  pages={3828--3850},
  year={2024}
}

@article{dekoninck2026beyondaime2025,
  title={Beyond benchmarks: Matharena as an evaluation platform for mathematics with llms},
  author={Dekoninck, Jasper and Jovanovi{\'c}, Nikola and Gehrunger, Tim and R{\"o}gnvaldsson, K{\'a}ri and Petrov, Ivo and Sun, Chenhao and Vechev, Martin},
  journal={arXiv preprint arXiv:2605.00674},
  year={2026}
}

@article{besag1974spatialpotts,
  title={Spatial interaction and the statistical analysis of lattice systems},
  author={Besag, Julian},
  journal={Journal of the Royal Statistical Society: Series B (Methodological)},
  volume={36},
  number={2},
  pages={192--225},
  year={1974},
  publisher={Wiley Online Library}
}

@inproceedings{potts1952somepotts,
  title={Some generalized order-disorder transformations},
  author={Potts, Renfrey Burnard},
  booktitle={Mathematical proceedings of the cambridge philosophical society},
  volume={48},
  number={1},
  pages={106--109},
  year={1952},
  organization={Cambridge University Press}
}

@article{zhang1992meanmrf,
  title={The mean field theory in EM procedures for Markov random fields},
  author={Zhang, Jun},
  journal={IEEE Transactions on signal processing},
  volume={40},
  number={10},
  pages={2570--2583},
  year={1992},
  publisher={IEEE}
}

@article{mcgrory2009variationalmrf,
  title={Variational Bayes for estimating the parameters of a hidden Potts model},
  author={McGrory, Clare A and Titterington, D Michael and Reeves, Robert and Pettitt, Anthony N},
  journal={Statistics and Computing},
  volume={19},
  number={3},
  pages={329--340},
  year={2009},
  publisher={Springer}
}

@article{sheng2024hybridflow,
  title   = {HybridFlow: A Flexible and Efficient RLHF Framework},
  author  = {Guangming Sheng and Chi Zhang and Zilingfeng Ye and Xibin Wu and Wang Zhang and Ru Zhang and Yanghua Peng and Haibin Lin and Chuan Wu},
  year    = {2024},
  journal = {arXiv preprint arXiv: 2409.19256}
}

@misc{yang2024qwen25mathtechnicalreportmathematical,
      title={Qwen2.5-Math Technical Report: Toward Mathematical Expert Model via Self-Improvement}, 
      author={An Yang and Beichen Zhang and Binyuan Hui and Bofei Gao and Bowen Yu and Chengpeng Li and Dayiheng Liu and Jianhong Tu and Jingren Zhou and Junyang Lin and Keming Lu and Mingfeng Xue and Runji Lin and Tianyu Liu and Xingzhang Ren and Zhenru Zhang},
      year={2024},
      eprint={2409.12122},
      archivePrefix={arXiv},
      primaryClass={cs.CL},
      url={https://arxiv.org/abs/2409.12122}, 
}

@misc{grattafiori2024llama3herdmodels,
      title={The Llama 3 Herd of Models}, 
      author={Aaron Grattafiori and Abhimanyu Dubey and Abhinav Jauhri and Abhinav Pandey and Abhishek Kadian and Ahmad Al-Dahle and Aiesha Letman and Akhil Mathur and Alan Schelten and Alex Vaughan and Amy Yang and Angela Fan and Anirudh Goyal and Anthony Hartshorn and Aobo Yang and others},
      year={2024},
      eprint={2407.21783},
      archivePrefix={arXiv},
      primaryClass={cs.AI},
      url={https://arxiv.org/abs/2407.21783}, 
}

@inproceedings{wu2026reasoning,
  title={Reasoning or memorization? unreliable results of reinforcement learning due to data contamination},
  author={Wu, Mingqi and Zhang, Zhihao and Dong, Qiaole and Xi, Zhiheng and Zhao, Jun and Jin, Senjie and Fan, Xiaoran and Zhou, Yuhao and Lv, Huijie and Zhang, Ming and others},
  booktitle={Proceedings of the AAAI Conference on Artificial Intelligence},
  volume={40},
  number={40},
  pages={33944--33952},
  year={2026}
}

@inproceedings{bengio2009icml-curriculum,
  title     = {{Curriculum Learning}},
  author    = {Bengio, Yoshua and Louradour, Jérôme and Collobert, Ronan and Weston, Jason},
  booktitle = {International Conference on Machine Learning},
  year      = {2009},
  pages     = {41-48},
  doi       = {10.1145/1553374.1553380},
  url       = {https://mlanthology.org/icml/2009/bengio2009icml-curriculum/}
}

@InProceedings{pmlr-v162-mindermann22a,
  title = 	 {Prioritized Training on Points that are Learnable, Worth Learning, and not yet Learnt},
  author =       {Mindermann, S{\"o}ren and Brauner, Jan M and Razzak, Muhammed T and Sharma, Mrinank and Kirsch, Andreas and Xu, Winnie and H{\"o}ltgen, Benedikt and Gomez, Aidan N and Morisot, Adrien and Farquhar, Sebastian and Gal, Yarin},
  booktitle = 	 {Proceedings of the 39th International Conference on Machine Learning},
  pages = 	 {15630--15649},
  year = 	 {2022},
  editor = 	 {Chaudhuri, Kamalika and Jegelka, Stefanie and Song, Le and Szepesvari, Csaba and Niu, Gang and Sabato, Sivan},
  volume = 	 {162},
  series = 	 {Proceedings of Machine Learning Research},
  month = 	 {17--23 Jul},
  publisher =    {PMLR},
  url = 	 {https://proceedings.mlr.press/v162/mindermann22a.html}
}

@inproceedings{bae2026online,
  title={Online difficulty filtering for reasoning oriented reinforcement learning},
  author={Bae, Sanghwan and Hong, Jiwoo and Lee, Min Young and Kim, Hanbyul and Nam, JeongYeon and Kwak, Donghyun},
  booktitle={Proceedings of the 19th Conference of the European Chapter of the Association for Computational Linguistics (Volume 1: Long Papers)},
  pages={700--719},
  year={2026}
}

@inproceedings{fang2026allocate,
  title={How to allocate, how to learn? dynamic rollout allocation and advantage modulation for policy optimization},
  author={Fang, Yangyi and Lin, Jiaye and Fu, Xiaoliang and Qin, Cong and Shi, Haolin and Hu, Chaowen and Pan, Lu and Zeng, Ke and Cai, Xunliang},
  booktitle={Findings of the Association for Computational Linguistics: ACL 2026},
  pages={14727--14744},
  year={2026}
}

@article{xiong2025reinforce,
  title={Reinforce-ada: An adaptive sampling framework for reinforce-style llm training},
  author={Xiong, Wei and Ye, Chenlu and Liao, Baohao and Dong, Hanze and Xu, Xinxing and Monz, Christof and Bian, Jiang and Jiang, Nan and Zhang, Tong},
  journal={arXiv e-prints},
  pages={arXiv--2510},
  year={2025}
}

@inproceedings{lalor-etal-2016-building,
    title = "Building an Evaluation Scale using Item Response Theory",
    author = "Lalor, John P.  and
      Wu, Hao  and
      Yu, Hong",
    editor = "Su, Jian  and
      Duh, Kevin  and
      Carreras, Xavier",
    booktitle = "Proceedings of the 2016 Conference on Empirical Methods in Natural Language Processing",
    month = nov,
    year = "2016",
    address = "Austin, Texas",
    publisher = "Association for Computational Linguistics",
    url = "https://aclanthology.org/D16-1062/",
    doi = "10.18653/v1/D16-1062",
    pages = "648--657"
}

@inproceedings{rodriguez-etal-2021-evaluation,
    title = "Evaluation Examples are not Equally Informative: How should that change {NLP} Leaderboards?",
    author = "Rodriguez, Pedro  and
      Barrow, Joe  and
      Hoyle, Alexander  and
      Lalor, John P.  and
      Jia, Robin  and
      Boyd-Graber, Jordan",
    editor = "Zong, Chengqing  and
      Xia, Fei  and
      Li, Wenjie  and
      Navigli, Roberto",
    booktitle = "Proceedings of the 59th Annual Meeting of the Association for Computational Linguistics and the 11th International Joint Conference on Natural Language Processing (Volume 1: Long Papers)",
    month = aug,
    year = "2021",
    address = "Online",
    publisher = "Association for Computational Linguistics",
    url = "https://aclanthology.org/2021.acl-long.346/",
    doi = "10.18653/v1/2021.acl-long.346",
    pages = "4486--4503"
}

@inproceedings{swayamdipta-etal-2020-dataset,
    title = "Dataset Cartography: Mapping and Diagnosing Datasets with Training Dynamics",
    author = "Swayamdipta, Swabha  and
      Schwartz, Roy  and
      Lourie, Nicholas  and
      Wang, Yizhong  and
      Hajishirzi, Hannaneh  and
      Smith, Noah A.  and
      Choi, Yejin",
    editor = "Webber, Bonnie  and
      Cohn, Trevor  and
      He, Yulan  and
      Liu, Yang",
    booktitle = "Proceedings of the 2020 Conference on Empirical Methods in Natural Language Processing (EMNLP)",
    month = nov,
    year = "2020",
    address = "Online",
    publisher = "Association for Computational Linguistics",
    url = "https://aclanthology.org/2020.emnlp-main.746/",
    doi = "10.18653/v1/2020.emnlp-main.746",
    pages = "9275--9293"
}

@inproceedings{zhu2003semi,
  title={Semi-supervised learning using gaussian fields and harmonic functions},
  author={Zhu, Xiaojin and Ghahramani, Zoubin and Lafferty, John D},
  booktitle={Proceedings of the 20th International conference on Machine learning (ICML-03)},
  pages={912--919},
  year={2003}
}

@inproceedings{NIPS2003_87682805,
 author = {Zhou, Dengyong and Bousquet, Olivier and Lal, Thomas and Weston, Jason and Sch\"{o}lkopf, Bernhard},
 booktitle = {Advances in Neural Information Processing Systems},
 editor = {S. Thrun and L. Saul and B. Sch\"{o}lkopf},
 pages = {},
 publisher = {MIT Press},
 title = {Learning with Local and Global Consistency},
 url = {https://proceedings.neurips.cc/paper_files/paper/2003/file/87682805257e619d49b8e0dfdc14affa-Paper.pdf},
 volume = {16},
 year = {2003}
}

@article{JMLR:v7:belkin06a,
  author  = {Mikhail Belkin and Partha Niyogi and Vikas Sindhwani},
  title   = {Manifold  Regularization: A Geometric Framework for Learning from Labeled and Unlabeled Examples},
  journal = {Journal of Machine Learning Research},
  year    = {2006},
  volume  = {7},
  number  = {85},
  pages   = {2399--2434},
  url     = {http://jmlr.org/papers/v7/belkin06a.html}
}

@article{geman1984stochastic,
  title={Stochastic relaxation, Gibbs distributions, and the Bayesian restoration of images},
  author={Geman, Stuart and Geman, Donald},
  journal={IEEE Transactions on pattern analysis and machine intelligence},
  number={6},
  pages={721--741},
  year={1984},
  publisher={IEEE}
}

@article{wainwright2008graphical,
  title={Graphical models, exponential families, and variational inference},
  author={Wainwright, Martin J and Jordan, Michael I},
  journal={Foundations and Trends{\textregistered} in Machine Learning},
  volume={1},
  number={1-2},
  pages={1--305},
  year={2008},
  publisher={Emerald Publishing Limited}
}

@article{Blei03042017,
author = {David M. Blei and Alp Kucukelbir and Jon D. McAuliffe},
title = {Variational Inference: A Review for Statisticians},
journal = {Journal of the American Statistical Association},
volume = {112},
number = {518},
pages = {859--877},
year = {2017},
publisher = {Taylor \& Francis},
doi = {10.1080/01621459.2017.1285773},
URL = { 
    
        https://doi.org/10.1080/01621459.2017.1285773
},
eprint = { 
        https://doi.org/10.1080/01621459.2017.1285773
        }
}

@article{hoffman2013stochastic,
  title={Stochastic variational inference},
  author={Hoffman, Matthew D and Blei, David M and Wang, Chong and Paisley, John},
  journal={Journal of machine learning research},
  year={2013}
}

@inproceedings{NIPS2013_51ef186e,
 author = {Broderick, Tamara and Boyd, Nicholas and Wibisono, Andre and Wilson, Ashia and Jordan, Michael},
 booktitle = {Advances in Neural Information Processing Systems},
 editor = {C.J. Burges and L. Bottou and M. Welling and Z. Ghahramani and K. Weinberger},
 pages = {},
 publisher = {Curran Associates, Inc.},
 title = {Streaming Variational Bayes},
 url = {https://proceedings.neurips.cc/paper_files/paper/2013/file/51ef186e18dc00c2d31982567235c559-Paper.pdf},
 volume = {26},
 year = {2013}
}

@article{ng2001spectral,
  title={On spectral clustering: Analysis and an algorithm},
  author={Ng, Andrew and Jordan, Michael and Weiss, Yair},
  journal={Advances in neural information processing systems},
  volume={14},
  year={2001}
}

@article{von2007tutorial,
  title={A tutorial on spectral clustering},
  author={Von Luxburg, Ulrike},
  journal={Statistics and computing},
  volume={17},
  number={4},
  pages={395--416},
  year={2007},
  publisher={Springer}
}

@misc{zhang2025qwen3embeddingadvancingtext,
      title={Qwen3 Embedding: Advancing Text Embedding and Reranking Through Foundation Models}, 
      author={Yanzhao Zhang and Mingxin Li and Dingkun Long and Xin Zhang and Huan Lin and Baosong Yang and Pengjun Xie and An Yang and Dayiheng Liu and Junyang Lin and Fei Huang and Jingren Zhou},
      year={2025},
      eprint={2506.05176},
      archivePrefix={arXiv},
      primaryClass={cs.CL},
      url={https://arxiv.org/abs/2506.05176}, 
}

@inproceedings{
wang2026scheduling,
title={Scheduling Your {LLM} Reinforcement Learning with Reasoning Trees},
author={Hong Wang and Zhezheng Hao and Jian Luo and Chenxing Wei and Yao Shu and Lei Liu and Cheaterlin and Hande Dong and Jiawei Chen},
booktitle={The Fourteenth International Conference on Learning Representations},
year={2026},
url={https://openreview.net/forum?id=V4zln7XiJj}
}

@misc{xu2026prunegenerateonlinerollout,
      title={Prune as You Generate: Online Rollout Pruning for Faster and Better RLVR}, 
      author={Haobo Xu and Sirui Chen and Ruizhong Qiu and Yuchen Yan and Chen Luo and Monica Cheng and Jingrui He and Hanghang Tong},
      year={2026},
      eprint={2603.24840},
      archivePrefix={arXiv},
      primaryClass={cs.CL},
      url={https://arxiv.org/abs/2603.24840}, 
}

@article{jain2024livecodebench,
    title={LiveCodeBench: Holistic and Contamination Free Evaluation of Large Language Models for Code},
    author={Jain, Naman and Han, King and Gu, Alex and Li, Wen-Ding and Yan, Fanjia and Zhang, Tianjun and Wang, Sida and Solar-Lezama, Armando and Sen, Koushik and Stoica, Ion},
    journal={arXiv preprint arXiv:2403.07974},
    year={2024}
}

@article{dixon1946sign,
  author  = {Dixon, Wilfrid J. and Mood, Alexander M.},
  title   = {The Statistical Sign Test},
  journal = {Journal of the American Statistical Association},
  volume  = {41},
  number  = {236},
  pages   = {557--566},
  year    = {1946},
  doi     = {10.1080/01621459.1946.10501898}
}

@article{pearson1900criterion,
  author  = {Pearson, Karl},
  title   = {On the Criterion That a Given System of Deviations from the
             Probable in the Case of a Correlated System of Variables Is Such
             That It Can Be Reasonably Supposed to Have Arisen from Random
             Sampling},
  journal = {The London, Edinburgh, and Dublin Philosophical Magazine and
             Journal of Science},
  volume  = {50},
  number  = {302},
  pages   = {157--175},
  year    = {1900},
  doi     = {10.1080/14786440009463897}
}

@book{cramer1946mathematical,
  author    = {Cram{\'e}r, Harald},
  title     = {Mathematical Methods of Statistics},
  publisher = {Princeton University Press},
  address   = {Princeton, NJ},
  year      = {1946}
}

@article{kruskal1952ranks,
  author  = {Kruskal, William H. and Wallis, W. Allen},
  title   = {Use of Ranks in One-Criterion Variance Analysis},
  journal = {Journal of the American Statistical Association},
  volume  = {47},
  number  = {260},
  pages   = {583--621},
  year    = {1952},
  doi     = {10.1080/01621459.1952.10483441}
}

@article{benjamini1995fdr,
  author  = {Benjamini, Yoav and Hochberg, Yosef},
  title   = {Controlling the False Discovery Rate: A Practical and Powerful
             Approach to Multiple Testing},
  journal = {Journal of the Royal Statistical Society: Series B
             (Methodological)},
  volume  = {57},
  number  = {1},
  pages   = {289--300},
  year    = {1995},
  doi     = {10.1111/j.2517-6161.1995.tb02031.x}
}

@misc{chen2024bgem3,
  title         = {{BGE M3-Embedding}: Multi-Lingual, Multi-Functionality,
                   Multi-Granularity Text Embeddings Through Self-Knowledge
                   Distillation},
  author        = {Chen, Jianlv and Xiao, Shitao and Zhang, Peitian and Luo,
                   Kun and Lian, Defu and Liu, Zheng},
  year          = {2024},
  eprint        = {2402.03216},
  archivePrefix = {arXiv},
  primaryClass  = {cs.CL},
  url           = {https://arxiv.org/abs/2402.03216}
}

@inproceedings{kusupati2022matryoshka,
  title     = {Matryoshka Representation Learning},
  author    = {Kusupati, Aditya and Bhatt, Gantavya and Rege, Aniket and
               Wallingford, Matthew and Sinha, Aditya and Ramanujan, Vivek and
               Howard-Snyder, William and Chen, Kaifeng and Kakade, Sham and
               Jain, Prateek and Farhadi, Ali},
  booktitle = {Advances in Neural Information Processing Systems},
  volume    = {35},
  year      = {2022},
  url       = {https://proceedings.neurips.cc/paper_files/paper/2022/hash/c32319f4868da7613d78af9993100e42-Abstract-Conference.html}
}

@inproceedings{reimers2019sentencebert,
  title     = {Sentence-{BERT}: Sentence Embeddings Using Siamese
               {BERT}-Networks},
  author    = {Reimers, Nils and Gurevych, Iryna},
  booktitle = {Proceedings of the 2019 Conference on Empirical Methods in
               Natural Language Processing and the 9th International Joint
               Conference on Natural Language Processing},
  pages     = {3982--3992},
  publisher = {Association for Computational Linguistics},
  year      = {2019},
  doi       = {10.18653/v1/D19-1410},
  url       = {https://aclanthology.org/D19-1410/}
}

\end{document}